%% file: main.tex
\documentclass[lettersize,journal]{IEEEtran}
\usepackage{amsmath,amsfonts}
\usepackage{algorithmic}
\usepackage{algorithm}
\usepackage{array}
\usepackage[caption=false,font=normalsize,labelfont=sf,textfont=sf]{subfig}
\usepackage{textcomp}
\usepackage{stfloats}
\usepackage{url}
\usepackage{verbatim}
\usepackage{graphicx}
\usepackage{cite}
\usepackage{colortbl}
\usepackage{xcolor}
\usepackage[pagebackref,breaklinks,colorlinks]{hyperref}

\newcommand{\etal}[0]{\textit{et al.}}

\begin{document}

\title{
Animatable and Relightable Gaussians\\for High-fidelity Human Avatar Modeling
}

\author{Zhe Li$^*$,
        Yipengjing Sun$^*$,
        Zerong Zheng,
        Lizhen Wang,
        Shengping Zhang
        and Yebin Liu,~\IEEEmembership{Member,~IEEE}
\\{\ttfamily \url{https://animatable-gaussians.github.io/relight}}
\thanks{$^*$ indicates equal contribution.}
\thanks{
Zhe Li, Lizhen Wang and Yebin Liu are with Department of Automation, Tsinghua University, Beijing 100084, P.R.China.
Yipengjing Sun and Shengping Zhang are with the School of Computer Science and Technology, Harbin Institute of Technology, Weihai 264209, P.R. China.
Zerong Zheng is with NNKosmos Technology, Hangzhou, P.R.China.
}%
\thanks{Corresponding author: Yebin Liu and Shengping Zhang.}
}

\markboth{Journal of \LaTeX\ Class Files,~Vol.~14, No.~8, August~2021}%
{Shell \MakeLowercase{\textit{et al.}}: A Sample Article Using IEEEtran.cls for IEEE Journals}

\maketitle

\input{secs/0_abstract}


\input{secs/1_introduction}

\input{secs/2_related_work}
\input{secs/3_method}
\input{secs/4_results}
\input{secs/5_discussion}

\bibliographystyle{IEEEtran}
\bibliography{ref}

\end{document}

%% file: secs/0_abstract.tex
\begin{abstract}
   Modeling animatable human avatars from RGB videos is a long-standing and challenging problem.
   Recent works usually adopt MLP-based neural radiance fields (NeRF) to represent 3D humans, but it remains difficult for pure MLPs to regress pose-dependent garment details.
   To this end, we introduce Animatable Gaussians, a new avatar representation that leverages powerful 2D CNNs and 3D Gaussian splatting to create high-fidelity avatars.
   To associate 3D Gaussians with the animatable avatar, we learn a parametric template from the input videos, and then parameterize the template on two front \& back canonical Gaussian maps where each pixel represents a 3D Gaussian.
   The learned template is adaptive to the wearing garments for modeling looser clothes like dresses.
   Such template-guided 2D parameterization enables us to employ a powerful StyleGAN-based CNN to learn the pose-dependent Gaussian maps for modeling detailed dynamic appearances.
   Furthermore, we introduce a pose projection strategy for better generalization given novel poses.
   To tackle the realistic relighting of animatable avatars, we introduce physically-based rendering into the avatar representation for decomposing avatar materials and environment illumination.
   Overall, our method can create lifelike avatars with dynamic, realistic, generalized and relightable appearances.
   Experiments show that our method outperforms other state-of-the-art approaches.
\end{abstract}

%% file: secs/1_introduction.tex
\begin{figure*}[t]
\hsize=\textwidth
\centering
\includegraphics[width=\linewidth]{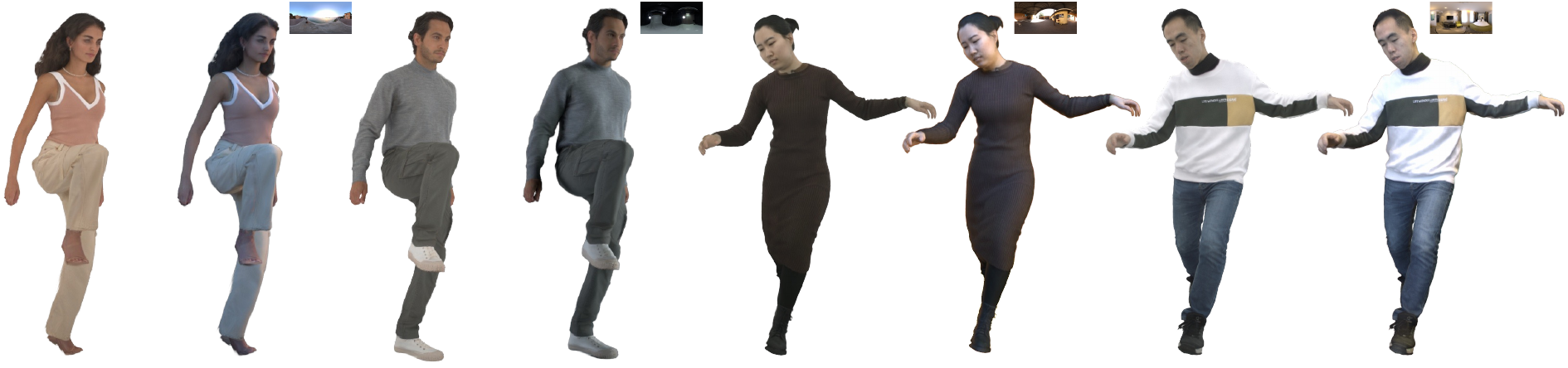}
\caption{
Lifelike relightable and animatable avatars with \textit{highly dynamic}, \textit{realistic} and \textit{generalized} details created by our method.
We show synthesized results animated by the same pose under the capture environment and novel lights.
}
\label{fig: teaser}
\end{figure*}

\section{Introduction}
\label{sec: intro}

\IEEEPARstart{A}{nimatable} human avatar modeling, due to its potential value in holoportation, Metaverse, game and movie industries, has been a popular topic in computer vision for decades.
However, how to effectively represent the human avatar is still a challenging problem.

Explicit representations, including both meshes and point clouds, are the prevailing choices, not just in human avatars but also throughout the entire 3D vision and graphics.
However, previous explicit avatar representations \cite{bagautdinov2021driving,xiang2022dressing,ma2021power} necessitate dense reconstructed meshes to model human geometry, thus limiting their applications in sparse-view video-based avatar modeling.
In the past few years, with the rise of implicit representations, particularly neural radiance fields (NeRF) \cite{mildenhall2020nerf}, many researchers tend to represent the 3D human as a pose-conditioned NeRF \cite{peng2021animatable,liu2021neural,zheng2022structured,li2022tava} to automatically learn a neural avatar from RGB videos.
However, implicit representations require a coordinate-based MLP to regress a continuous field, suffering from the low-frequency spectral bias \cite{tancik2020fourier} of MLPs.
Although many works aim to enhance the avatar representation by texture feature \cite{liu2021neural} or structured local NeRFs \cite{zheng2022structured}, they fail to produce satisfactory results because they still rely on an MLP to output the continuous implicit fields.

Recently, 3D Gaussian splatting \cite{kerbl2023gaussian}, an explicit and efficient point-based representation, has been proposed for both high-fidelity rendering quality and real-time rendering speed.
In contrary to implicit representations, explicit point-based representations have the potential to be parameterized on 2D maps \cite{ma2021power}, thus enabling us to employ more powerful 2D networks for modeling higher-fidelity avatars.
Based on this observation, we present \textit{Animatable Gaussians}, a new avatar representation that leverages 3D Gaussian splatting and powerful 2D CNNs for realistic avatar modeling.
The first challenge lies in modeling general garments including long dresses.
Inspired by point-based geometric avatars \cite{lin2022learning,ma2022neural}, we first reconstruct a parametric template from the input videos and inherit the parameters of SMPL \cite{loper2015smpl} by diffusing the skinning weights \cite{lin2022learning}.
The character-specific template models the basic shapes of the wearing garments, even for long dresses. 
This allows us to animate 3D Gaussians in accordance with the template motion while avoiding density control in standard Gaussians \cite{kerbl2023gaussian}, thereby ensuring the maintenance of a temporally consistent structure for 3D Gaussians in the following 2D parameterization.

For compatibility with 2D networks, it is necessary to parameterize the 3D template onto 2D maps.
However, it remains challenging to unwrap the template with arbitrary topologies onto a unified and continuous UV space.
Regarding that the front \& back views almost cover the entire canonical human, we achieve the parameterization by orthogonally projecting the canonical template to both views.
In each view, we define every pixel within the template mask as a 3D Gaussian, represented by its position, covariance, opacity, and color attributes, resulting in two front \& back Gaussian maps.
Similarly, given the driving pose, we obtain two posed position maps that serve as the pose conditions.
Such a template-guided parameterization enables predicting pose-dependent Gaussian maps from the pose conditions through a powerful StyleGAN-based \cite{karras2019style,karras2020analyzing,karras2021alias} conditional generator, StyleUNet \cite{wang2023styleavatar}.

Benefiting from the powerful 2D CNNs and explicit 3D Gaussian splatting, our method can faithfully reconstruct human details under training poses.
On the other hand, given novel poses, the generalization of animatable avatars has not been extensively explored.
Due to the data-driven nature of learning-based avatar modeling, direct extrapolation to poses out of distribution will certainly yield unsatisfactory results.
Therefore, we propose to employ Principal Component Analysis (PCA) to project the driving pose signal, represented by the position maps, into the PCA space, facilitating reasonable interpolation within the distribution of training poses.
Such a pose projection strategy realizes reasonable and high-quality synthesis for novel poses.

A preliminary version of this work has been published in CVPR 2024 \cite{li2024animatablegaussians}, in which we propose a novel avatar representation for modeling realistic animatable human avatars from multi-view videos.
However, the preliminary work \cite{li2024animatablegaussians} can only animate the avatar under illumination from the capture environment.
In the current version, we further introduce physically-based rendering (PBR) \cite{jin2023tensoir,gao2023relightable} into our avatar representation for creating both animatable and relightable human avatars.
Specifically, besides the original attributes of 3D Gaussians, our model also predicts albedo, roughness, and light visibility to decompose the avatar materials and illumination of the capture environment.
As a result, our model can produce realistic animation under novel illumination.

We extend the preliminary version \cite{li2024animatablegaussians} in the following ways.
First, we introduce PBR into our avatar representation (Animatable Gaussians) for creating relightable human avatars.
Second, we compare our method with concurrent works on 3D Gaussian splatting-based avatars, and our method outperforms them on the avatar quality.
Third, we compare our method with other works on human performance relighting, and our method can produce more realistic human relighting.

In summary, our technical contributions are:
\begin{itemize}
    \item Animatable Gaussians, a new avatar representation that introduces explicit 3D Gaussian splatting into avatar modeling to employ powerful 2D CNNs for creating lifelike avatars with high-fidelity pose-dependent dynamics.
    \item Template-guided parameterization that learns a character-specific template for general clothes like dresses, and parameterizes 3D Gaussians onto front \& back Gaussian maps for compatibility with 2D networks.
    \item A simple yet effective pose projection strategy that employs PCA on the driving signal, promoting better generalization to novel poses.
    \item We introduce physically-based rendering into Animatable Gaussians for photorealistic relighting under novel illumination.
\end{itemize}

Overall, benefiting from these contributions, our method can create lifelike animatable and relightable avatars with \textit{highly dynamic}, \textit{realistic} and \textit{generalized} appearances as shown in Fig.~\ref{fig: teaser}.
The code is available at \url{https://github.com/lizhe00/AnimatableGaussians}, and earns more than 700 stars.

%% file: secs/2_related_work.tex
\section{Related Work}
\label{sec: related work}

\subsection{Mesh-based Human Avatars}
\label{subsec: mesh avatars}
The polygon mesh is the most popular 3D representation for its compatibility with traditional rendering pipelines.
To model animatable human avatars using meshes, early approaches propose to reconstruct a character-specific textured mesh and animate it by physical simulation \cite{stoll2010video,guan2012drape} or retrieval from a database \cite{xu2011video}.
Recently, researchers tend to utilize neural networks to model dynamic textures and motions.
Bagautdinov \etal \cite{bagautdinov2021driving}, Xiang \etal \cite{xiang2021modeling,xiang2022dressing} and Halimi \etal \cite{halimi2022pattern} reconstruct topology-consistent meshes from dense multi-view videos and learn the dynamic texture in a UV space.
DDC \cite{habermann2021real} and HDHumans \cite{habermann2023hdhumans} learn the deformation parameterized by both skeletons and embedded graph \cite{sumner2007embedded} of a pre-scanned template.
DELIFFAS \cite{Kwon2023deliffas} employs DDC as a deformable template and parameterizes the light field around the body onto double surfaces for fast synthesis.
These mesh-based methods require dense reconstruction, non-rigid tracking, or pre-scanned templates for representing dynamic humans.
Besides, some works optimize the non-rigid deformation upon SMPL \cite{loper2015smpl} from a monocular RGB \cite{alldieck2018video,alldieck2018detailed,zhao2022high} or RGB-D \cite{burov2021dynamic,kim2022laplacianfusion} video, but the avatar quality is limited by the SMPL+D representation.

\subsection{Implicit Function-based Human Avatars}
\label{subsec: implicit function avatars}
Implicit function is a coordinate-based function, usually represented by an MLP, that outputs a continuous field, e.g., signed distance function (SDF) \cite{jiang2022selfrecon,xu2023relightable}, occupancy \cite{mihajlovic2021leap}, and radiance (NeRF) \cite{peng2021animatable} fields.
In geometric avatar modeling, many works represent the human avatar as pose-conditioned SDF \cite{saito2021scanimate,wang2021metaavatar,tiwari2021neural,dong2022pina,ho2023learning} or occupancy \cite{deng2020nasa,chen2021snarf,chen2023fast,mihajlovic2021leap,mihajlovic2022coap,li2022avatarcap} fields learned from human scans or depth sequences.
In contrast, NeRF containing a density and color field is widely used in textured avatar modeling \cite{peng2022animatable,su2021a-nerf,Feng2022scarf,weng2022humannerf,te2022neural,peng2022selfnerf,guo2023vid2avatar,jiang2023instantavatar,jiang2022neuman} because of its good differentiable property.
Animatable NeRF \cite{peng2021animatable} introduces SMPL deformation into NeRF for animatable human modeling.
Neural Actor \cite{liu2021neural} and UV volumes \cite{chen2023uv} parameterize 3D humans on SMPL or DensePose \cite{guler2018densepose} UV space, thus limiting modeling loose clothes far from the human body.
SLRF \cite{zheng2022structured} defines local NeRF around sampled nodes upon SMPL and learns the pose-dependent dynamics in the local space.
TAVA \cite{li2022tava} models the human or animal deformation using only 3D skeletons without the requirement of a parametric model.
ARAH \cite{wang2022arah} represents the avatar geometry as SDF and adopts SDF-based volume rendering \cite{yariv2021volume,wang2021neus} for learning more plausible geometry from RGB videos.
DANBO \cite{su2022danbo} employs GNNs to learn the part-based pose feature.
Li \etal \cite{li2023posevocab} introduce a learnable pose vocabulary to learn higher-frequency pose conditions for the conditional NeRF.
Besides the body avatar, TotalSelfScan \cite{dong2022totalselfscan}, X-Avatar \cite{shen2023xavatar} and AvatarReX \cite{zheng2023avatarrex} propose compositional full-body avatars for expressive control of the human body, hands and face.
However, the implicit function-based methods usually adopt pure MLPs to represent the human avatar, yielding smooth or blurry quality due to the low-frequency bias of MLPs \cite{tancik2020fourier}.
What's worse, the rendering speed of these methods is usually slow because rendering from implicit fields requires dense sampling along a ray.

\subsection{Point-based Human Avatars}
\label{subsec: point avatars}
Point cloud is also a powerful and popular representation in human avatar modeling.
Given 3D scans of a character, SCALE \cite{ma2021scale} and POP \cite{ma2021power} learn the non-rigid deformation of dense points on SMPL UV maps to represent the dynamic garment wrinkles.
FITE \cite{lin2022learning} and CloSET \cite{zhang2023closet} extract pose features from projective maps or PointNet \cite{qi2017pointnet,qi2017pointnet++} to avoid discontinuity on the UV map.
SKiRT \cite{ma2022neural} and FITE learn a coarse template from the input scans and utilize learned or diffused skinning weights to animate loose clothes.
Prokudin \etal \cite{prokudin2023dynamic} propose dynamic point fields for general dynamic reconstruction.
This work and NPC \cite{su2023npc} show results on avatars created from RGB videos using Point-NeRF \cite{xu2022pointnerf}.
However, applying Point-NeRF to avatar modeling still relies on a low-frequency coordinate-based MLP, struggling with the same problems in Sec.~\ref{subsec: implicit function avatars}.
On the other hand, point-based rendering via splatting \cite{pfister2000surfels,zwicker2001surface,zwicker2002pointshop,zwicker2004perspective,yifan2019differentiable,aliev2020neural,lassner2021pulsar,kopanas2021point,ruckert2022adop} offers another probability for animatable avatar modeling.
PointAvatar \cite{zheng2023pointavatar} learns a canonical point cloud and deformation field to model head avatars from a monocular video via PyTorch3D’s \cite{ravi2020accelerating} differentiable point renderer.

Recently, 3D Gaussian splatting \cite{kerbl2023gaussian} (3DGS), an efficient differentiable point-based rendering method, has been proposed for real-time photo-realistic scene rendering.
Along with extending 3DGS to dynamic scene modeling \cite{luiten2023dynamic,yang2023deformable,wu20234d,yang2023real}, many researchers also introduced 3DGS into animatable human avatars.
Specifically, D3GA \cite{zielonka2023drivable} leverages cage-based deformation to model the motion of 3D Gaussians.
While other approaches like GART\cite{lei2023gart}, 3DGS-Avatar \cite{qian20233dgs}, GauHuman \cite{hu2023gauhuman} and HUGS \cite{kocabas2023hugs} employ linear blend skinning (LBS) to model human motions and reconstruct a 3DGS-based animatable avatar from monocular videos.
However, these approaches cannot produce highly realistic and dynamic human appearances because they all represent the canonical 3D human using MLPs, facing the same low-frequency problem as NeRF-based approaches (Sec.~\ref{subsec: implicit function avatars}).
We observe such an explicit point-based representation can be combined with 2D CNNs for high-quality avatar modeling.
Concurrent works including ASH \cite{pang2023ash} and GaussianAvatar \cite{hu2023gaussianavatar} parameterize the 3D character on a 2D UV map to predict Gaussian attributes using 2D CNNs, sharing similar ideas with our method.
Differently, we parameterize the canonical 3D human on the front and back views by orthographic projection.

\subsection{Human Relighting}
Human relighting aims to manipulate the reflectance field of the human surface, thus enabling an immersive fusion with novel illumination.
Conventional approaches \cite{chabert2006relighting, debevec2012light, debevec2000acquiring, debevec2002lighting, guo2019relightables, hawkins2001photometric, wenger2005performance, weyrich2006analysis} propose capturing the reflectance characteristics of a human subject through a LightStage arrangement. This setup entails controlled illumination systems and dense camera arrays, facilitating the generation of photorealistic renderings under diverse lighting conditions. However, such configurations are both financially demanding to capture and not readily accessible to the public. With the advancement of neural implicit representations, recent methods~\cite{chen2022relighting4d, iqbal2023rana, sun2023neural, xu2023relightable, lin2024relightable, wang2023intrinsicavatar} necessitate solely multi-view or even monocular video recordings obtained under constant unknown illumination conditions to model both human motion and light transport properties. Relighting4D \cite{chen2022relighting4d} employs NeuralBody~\cite{peng2021neural} as the dynamic human model and utilizes neural inverse rendering to decompose it into a 3D reflectance field, enabling the estimation of materials and lighting properties. Building upon this framework, Sun \etal \cite{sun2023neural} apply LBS to transform the reflectance field from the canonical space to the observation space, facilitating animated sequences with relighting effects. However, this work does not address shadowing effects. RANA \cite{iqbal2023rana} pretrains an SMPL+D-based representation incorporating albedo and normal map refinement techniques. It utilizes a simplified spherical harmonics lighting model to achieve relighting effects but lacks the capability to accurately model specular effects and shadows. Lin \etal \cite{lin2024relightable} propose to estimate pose-aware light visibility through part-wise MLPs, demonstrating improved shadowing estimation when generalizing to unseen poses. Xu \etal \cite{xu2023relightable}  utilize Hierarchical Distance Queries (HDQ) via sphere tracing to calculate correct SDF values under arbitrary human poses and then incorporate distant field soft shadow (DFSS) for estimating reasonable soft visibility maps. IntrinsicAvatar \cite{wang2023intrinsicavatar} employs explicit Monte-Carlo ray tracing in canonical space to capture secondary shading effects, thereby enabling precise estimation of materials and environmental lighting. However, most previous methods rely on NeRF-based techniques, which inherently yield limited rendering quality. Thanks to the proposed powerful Gaussian representation, our method can achieve more vivid and realistic relighting results under novel poses. 

%% file: secs/3_method.tex
\begin{figure*}[t]
    \centering
    \includegraphics[width=\linewidth]{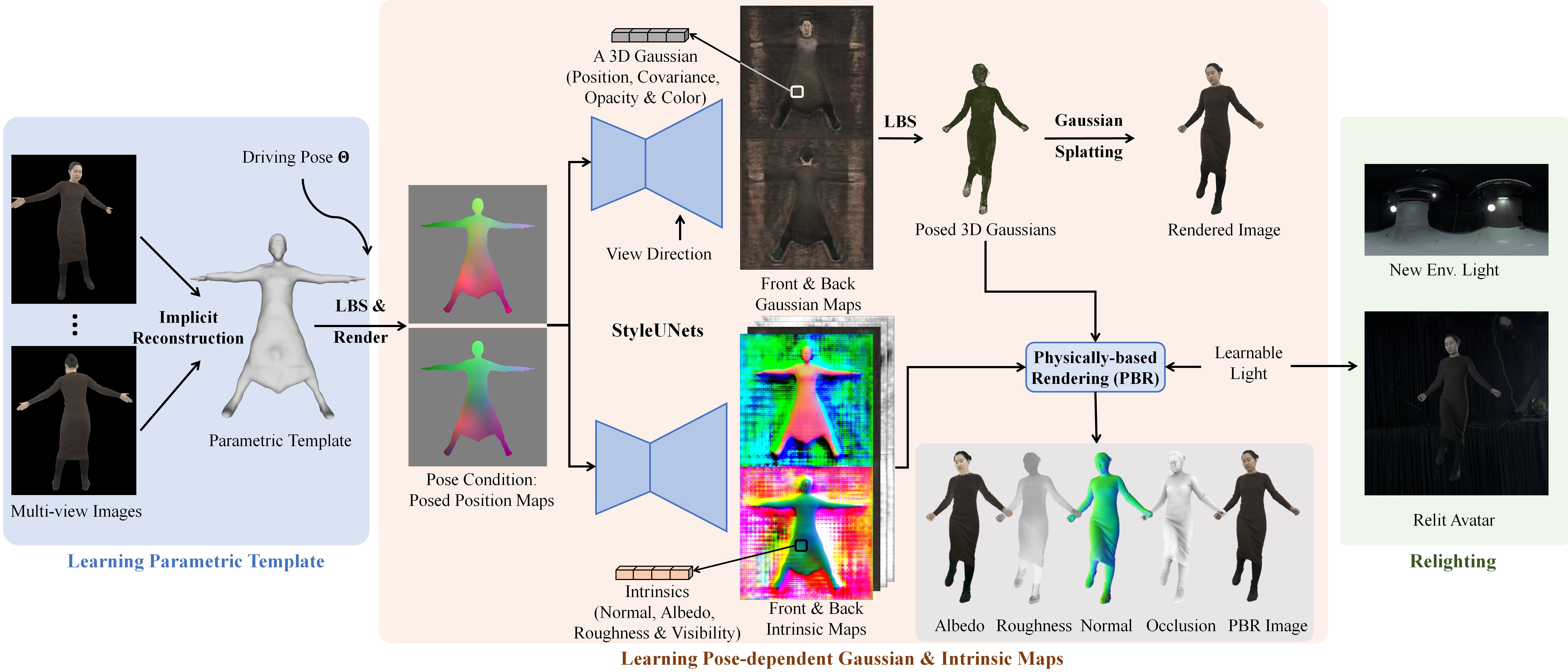}
    \caption{
    \textbf{Illustration of the avatar modeling pipeline.} 
    It contains two main steps: 1) Reconstruct a character-specific template from multi-view images. 
    2) Predict pose-dependent Gaussian and intrinsic maps through StyleUNets, and render the posed Gaussians by Gaussian splatting and physically-based rendering to learn both pose-dependent dynamics and avatar materials.
    Finally, given a novel environment light, we can animate the avatar with realistic dynamic appearances and shadow effects.
    }
    \label{fig: overview}
\end{figure*}

\section{Method}
\label{sec: method}

\subsection{Preliminary: 3D Gaussian Splatting}
3D Gaussian splatting \cite{kerbl2023gaussian} is an explicit point-based 3D representation that consists of a set of 3D Gaussians.
Each 3D Gaussian is parameterized by its position (mean) $\boldsymbol{\mu}$, covariance matrix $\mathbf{\Sigma}$, opacity $\alpha$ and color $\mathbf{c}$, and its probability density function is formulated as
\begin{equation}
    f(\mathbf{x}|\boldsymbol{\mu},\mathbf{\Sigma})=\exp\left(-\frac{1}{2} (\mathbf{x} - \boldsymbol{\mu})^\top \mathbf{\Sigma}^{-1} (\mathbf{x} - \boldsymbol{\mu})\right),
    \label{eq: gaussian pdf}
\end{equation}
where we omit the constant factor in Eq.~\ref{eq: gaussian pdf}.
For rendering a 2D image, the 3D Gaussians are splatted onto 2D planes, resulting in 2D Gaussians.
The pixel color $\mathbf{C}$ is computed by blending $N$ ordered 2D Gaussians overlapping this pixel:
\begin{equation}
    \mathbf{C} = \sum_{i=1}^N \alpha_i \prod_{j=1}^{i-1}(1-\alpha_j)\mathbf{c}_i,
    \label{eq: rasterization}
\end{equation}
where $\mathbf{c}_i$ is the color of each 2D Gaussian, and $\alpha_i$ is the blending weight derived from the learned opacity and 2D Gaussian distribution \cite{zwicker2001surface}.

\subsection{Overview}
\label{subsec: overview}
Given multi-view RGB videos of a character and the corresponding SMPL-X \cite{SMPL-X:2019} registrations about the per-frame pose and shared shape, our objective is to create a lifelike animatable avatar.
As illustrated in Fig.~\ref{fig: overview}, our method contains two main steps:
\begin{enumerate}
    \item \textbf{Learning Parametric Template.}
    We begin by selecting a frame with a near A-pose from the input videos, and then optimize a canonical SDF and color field to fit the multi-view images through SMPL skinning and SDF-based volume rendering \cite{yariv2021volume}.
    The template mesh is subsequently extracted from the canonical SDF field using Marching Cubes \cite{lorensen1987marching}.
    We then diffuse the skinning weights from the SMPL vertices to the template surface, obtaining a deformable parametric template.
    \item \textbf{Learning Pose-dependent Gaussian and Intrinsic Maps.}
    Given a training pose, we first deform the template to the posed space via linear blend skinning (LBS) and render the posed vertex coordinates to canonical front \& back views to obtain two position maps.
    The position maps serve as the pose condition and are translated into front \& back Gaussian and intrinsic maps through StyleUNets \cite{wang2023styleavatar}.
    We then extract valid 3D Gaussians inside the template mask from the Gaussian map, and deform the canonical 3D Gaussians to the posed space by LBS.
    In one branch, we directly render the posed 3D Gaussians to a camera view using the observed colors on the Gaussian maps. 
    On the other hand, we render the posed 3D Gaussians using the PBR color computed from the intrinsic maps. 
    These two branches allow our method not only to learn pose-dependent animation but also to decompose the avatar appearance into materials and light conditions for relighting purposes.
    
\end{enumerate}

\subsection{Avatar Representation} 

\subsubsection{Learning Parametric Template}
Given the multi-view videos, we first select one frame in which the character is under a near A-pose. 
Our goal is to reconstruct a canonical geometric model as the template from the multi-view images.
Specifically, we represent the canonical character as an SDF and color field instantiated by an MLP.
To associate the canonical and posed spaces, we precompute a skinning weight volume $\mathcal{W}$ in the canonical space by diffusing the weights from the SMPL surface throughout the whole 3D volume along the surface normal \cite{lin2022learning}.
For each point in the posed space, we search its canonical correspondence by root finding \cite{chen2021snarf}:
\begin{equation}
    \min_{\mathbf{x}_c} \left\|\text{LBS}(\mathbf{x}_\text{c};\mathbf{\Theta},\mathcal{W}) - \mathbf{x}_\text{p}\right\|_2^2,
    \label{eq: root finding}
\end{equation}
where $\text{LBS}(\cdot)$ is a linear blend skinning function that transforms a canonical point $\mathbf{x}_\text{c}$ to its posed position $\mathbf{x}_\text{p}$ in accordance with the SMPL pose $\mathbf{\Theta}$.
Then the canonical correspondence is fed into the MLP to query its SDF and color, which are used to render RGB images by SDF-based volume rendering \cite{yariv2021volume}.
The rendered images are compared with the ground truth for optimizing the canonical fields via differentiable volume rendering.
Finally, we extract the geometric template from the SDF field and query the skinning weights for each vertex in the precomputed weight volume $\mathcal{W}$, obtaining a deformable parametric template.

\begin{figure}[t]
    \centering
    \includegraphics[width=\linewidth]{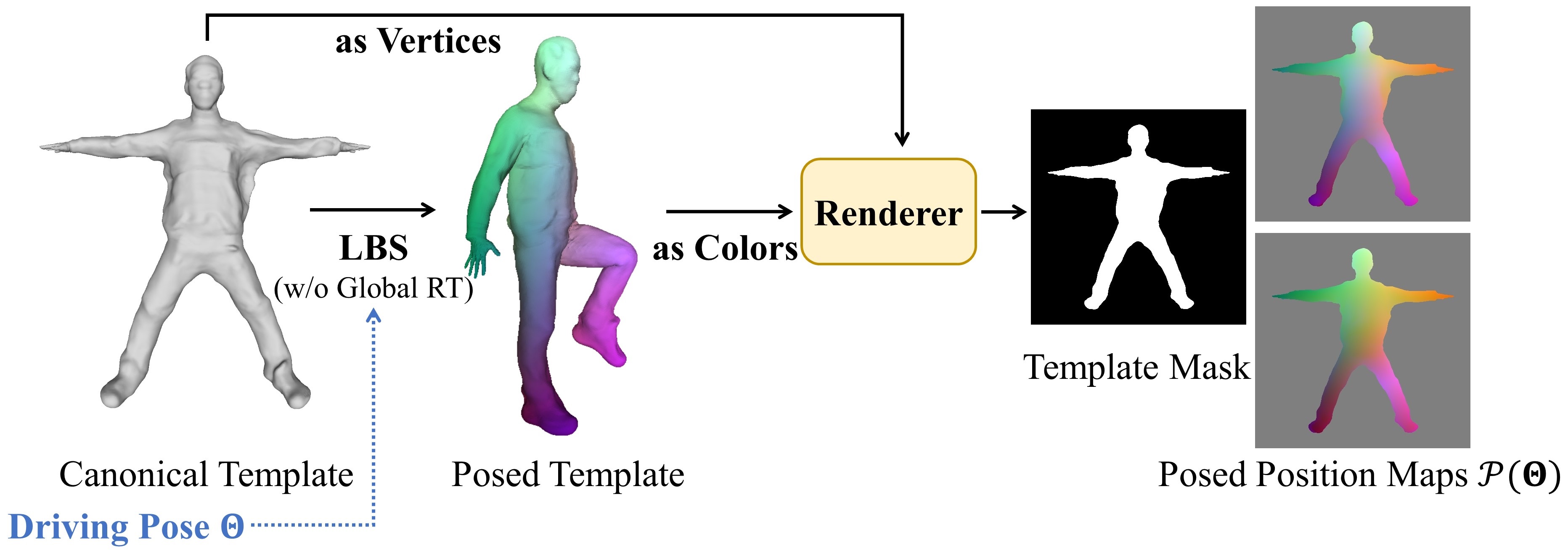}
    \caption{\textbf{Illustration of the posed position maps.}}
    \label{fig: pose_cond}
\end{figure}

\subsubsection{Template-guided Parameterization}
Previous human avatar representations in NeRF-based approaches \cite{peng2021animatable,liu2021neural,zheng2022structured} necessitate the coordinate-based MLPs for the formulation of the implicit NeRF function. 
However, MLPs have demonstrated a low-frequency bias \cite{tancik2020fourier}, hindering their ability to model high-frequency human dynamics.
In light of this observation, we replace MLPs with more powerful 2D CNNs for creating higher-quality human avatars.
To ensure compatibility with 2D networks, the 3D representation of the human avatar needs to be parameterized in 2D space.
Therefore, we propose to parameterize the 3D Gaussians anchored on the canonical template onto front \& back views via orthogonal projection.
As illustrated in Fig.~\ref{fig: pose_cond}, given a driving pose $\mathbf{\Theta}$, we first deform the template to the posed space via LBS.
Note that we do not consider the global transformation in this skinning process, because the global orientation and translation would not change the human dynamic details.
Then we take the posed coordinate as the vertex color on the canonical template, and render it to both front \& back views by orthogonal projection, obtaining posed position maps $\mathcal{P}_\text{f}(\mathbf{\Theta})$ and $\mathcal{P}_\text{b}(\mathbf{\Theta})$ that serve as pose conditions for the network.

\subsubsection{Pose-dependent Gaussian Maps}
\begin{figure}[t]
    \centering
    \includegraphics[width=0.8\linewidth]{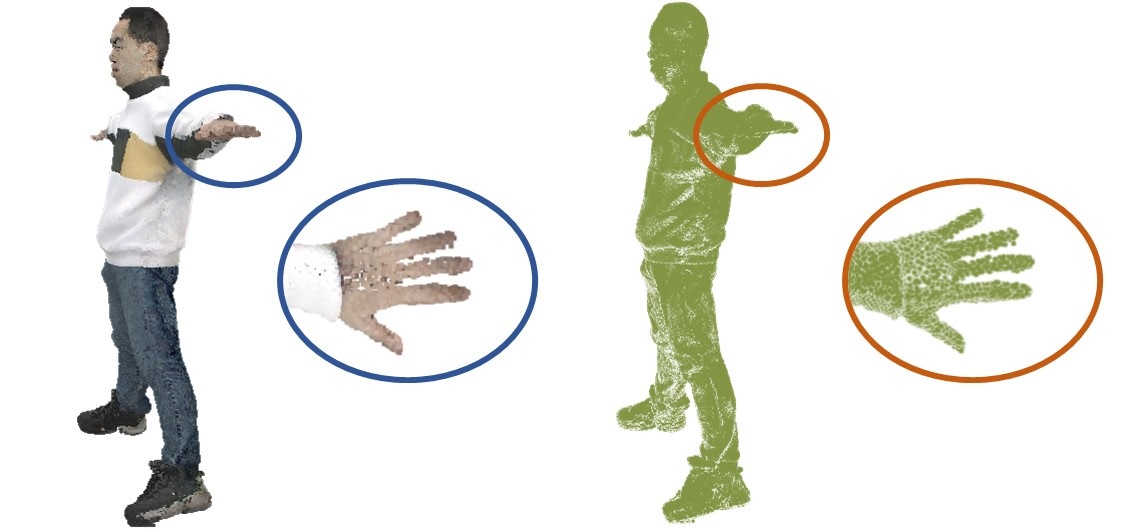}
    \caption{\textbf{Canonical 3D Gaussians on side regions and hands.}}
    \label{fig: side regions}
\end{figure}

We employ a powerful StyleGAN-based CNN , StyleUNet \cite{wang2023styleavatar} $\mathcal{F}_{\mathcal{G}}$, to predict pose-dependent Gaussian maps from the pose conditions:
\begin{equation}
    \mathcal{G}_\text{f}(\mathbf{\Theta}),\mathcal{G}_\text{b}(\mathbf{\Theta}) \leftarrow \mathcal{F}_{\mathcal{G}}(\mathcal{P}_\text{f}(\mathbf{\Theta}),\mathcal{P}_\text{b}(\mathbf{\Theta}),\mathcal{V}),
    \label{eq: styleunet}
\end{equation}
where $\mathcal{G}_\text{f}(\mathbf{\Theta})$ and $\mathcal{G}_\text{b}(\mathbf{\Theta})$ are front and back pose-dependent Gaussian maps, respectively, and each pixel represents a 3D Gaussian \cite{kerbl2023gaussian} including a position, covariance, opacity and color.
To ensure that the position attribute of predicted Gaussian maps approximates the canonical human body, we opt to predict an offset map $\Delta\mathcal{O}(\mathbf{\Theta})$ on the parametric template instead of a global position map.
We also modulate the output color attributes on Gaussian maps with a view direction map $\mathcal{V}$ to model view-dependent variance like NeRF-based approaches \cite{peng2021animatable}.
We extract canonical 3D Gaussians inside the template mask from the pose-dependent Gaussian maps.
It is worth mentioning that despite utilizing only front and back views for parameterizing the 3D Gaussians, the resulting point clouds still cover the side regions and hands of the human body as demonstrated in Fig.~\ref{fig: side regions}.
The reason is that the projection to front \& back views is orthographic, thus there exist sufficient 3D Gaussians to model these parts.

\subsubsection{LBS of 3D Gaussians}
To render the synthesized avatar under the driving pose, we need to deform the canonical 3D Gaussians to the posed space.
Specifically, given a canonical 3D Gaussian, we transform its position $\mathbf{p}_\text{c}$ and covariance $\mathbf{\Sigma}_\text{c}$ attributes:
\begin{equation}
\begin{split}
    \mathbf{p}_\text{p} = \mathbf{R} \mathbf{p}_\text{c} + \mathbf{t},\\
    \mathbf{\Sigma}_\text{p} = \mathbf{R} \mathbf{\Sigma}_\text{c} \mathbf{R}^\top,
\end{split}
\end{equation}
where $\mathbf{R}$ and $\mathbf{t}$ are the rotation matrix and translation vector calculated with the skinning weights of each 3D Gaussian.
Finally, we render the posed 3D Gaussians to a desired camera view through splatting-based rasterization (Eq.~\ref{eq: rasterization}).

\subsubsection{Pose-dependent Intrinsic Maps}
The pose-dependent Gaussian maps only represent the human appearances under the illumination of the capture environment, limiting animation under novel lighting conditions.
To disentangle the avatar geometry, material and lighting conditions, we leverage the classic rendering equation \cite{kajiya1986rendering} to simulate the rendering process:
\begin{equation}
    L_o(\mathbf{x},\boldsymbol{\omega}_o)=\int_\Omega L_i(\mathbf{x},\boldsymbol{\omega}) f(\mathbf{x},\boldsymbol{\omega}_i,\boldsymbol{\omega}_o;\alpha,\gamma,\mathbf{n})(\boldsymbol{\omega}_i\cdot \mathbf{n})d\boldsymbol{\omega}_i,
    \label{eq: pbr}
\end{equation}
where $L_o(\mathbf{x},\boldsymbol{\omega}_o)$ and $L_i(\mathbf{x},\boldsymbol{\omega}_i)$ are the outgoing and incident radiance at a surface position $\mathbf{x}$ along direction $\boldsymbol{\omega}_o$ and $\boldsymbol{\omega}_i$, respectively, and $\mathbf{n}$ is the normal vector at $\mathbf{x}$. $f(\mathbf{x},\boldsymbol{\omega}_i,\boldsymbol{\omega}_o;\alpha,\gamma,\mathbf{n})$ is the Bidirectional Reflectance Distribution Function
(BRDF) determined by the surface geometry (normal $\mathbf{n}$) and material properties including albedo $\alpha$ and roughness $\gamma$.
Considering the light visibility, the incident radiance is further formulated as
\begin{equation}
    L_\text{i}(\mathbf{x},\boldsymbol{\omega}_i)=V(\mathbf{x},\boldsymbol{\omega}_i)L_i(\boldsymbol{\omega}_i),
\end{equation}
where $V(\mathbf{x},\boldsymbol{\omega}_i)\in\{0,1\}$ indicates whether $\mathbf{x}$ is visible along the light direction $\boldsymbol{\omega}_i$. The global light $L_i(\boldsymbol{\omega}_i)$ is parametrized as learnable Spherical Harmonics (SH), which are optimized during the inverse rendering process.

Based on the physically-based rendering process, we additionally learn pose-dependent intrinsic maps including normal $\mathcal{N}(\mathbf{\Theta})$, albedo $\mathcal{A}(\mathbf{\Theta})$, and roughness $\boldsymbol{\gamma}(\mathbf{\Theta})$ maps on the front and back canonical views:
\begin{equation}
\begin{split}
    \mathcal{N}_\text{f}(\mathbf{\Theta}),\mathcal{N}_\text{b}(\mathbf{\Theta})\leftarrow\mathcal{F}_{\mathcal{N}}(\mathcal{P}_\text{f}(\mathbf{\Theta}),\mathcal{P}_\text{b}(\mathbf{\Theta})),\\
    \mathcal{A}_\text{f}(\mathbf{\Theta}),\mathcal{A}_\text{b}(\mathbf{\Theta})\leftarrow\mathcal{F}_{\mathcal{A}}(\mathcal{P}_\text{f}(\mathbf{\Theta}),\mathcal{P}_\text{b}(\mathbf{\Theta})),\\
    \boldsymbol{\gamma}_\text{f}(\mathbf{\Theta}),\boldsymbol{\gamma}_\text{b}(\mathbf{\Theta})\leftarrow\mathcal{F}_{\boldsymbol{\gamma}}(\mathcal{P}_\text{f}(\mathbf{\Theta}),\mathcal{P}_\text{b}(\mathbf{\Theta})),
\end{split}
\end{equation}
where $\mathcal{F}_{\mathcal{N}}$, $\mathcal{F}_{\mathcal{A}}$ and $\mathcal{F}_{\boldsymbol{\gamma}}$ are StyleUNet modules.
More specifically, the learned normal map $\mathcal{N}(\mathbf{\Theta})$ is an offset of the surface normal on the parametric template for better pose generalization.
The light visibility $V(\mathbf{x},\boldsymbol{\omega}_i)$ can be computed from the posed 3D Gaussians by point-based ray tracing \cite{gao2023relightable}.
However, as mentioned in \cite{gao2023relightable}, computing the light visibility during training is not preferred because of the computational complexity.
Therefore, we train an additional network to predict light visibility and supervise the prediction by randomly sampling view directions.
Specifically, we formulate the light-direction-dependent visibility as SH, and predict SH coefficient map $\mathcal{K}(\mathbf{\Theta})$ using a StyleUNet $\mathcal{F}_{\mathcal{K}}$:
\begin{equation}
    \mathcal{K}_\text{f}(\mathbf{\Theta}),\mathcal{K}_\text{b}(\mathbf{\Theta})\leftarrow\mathcal{F}_{\mathcal{K}}(\mathcal{P}_\text{f}(\mathbf{\Theta}),\mathcal{P}_\text{b}(\mathbf{\Theta})).
\end{equation}

Similar to the Gaussian maps, each pixel value on normal, albedo, roughness and light visibility maps is associated with a 3D Gaussian.
Following Relightable 3D Gaussian \cite{gao2023relightable}, we sample $N$ incident light directions over the hemisphere space, and calculate the PBR color of each 3D Gaussian by the discrete form of Eq.~\ref{eq: pbr}:
\begin{equation}
\begin{split}
&L_o(\boldsymbol{\mu},\boldsymbol{\omega}_o)=\\&\sum_{\boldsymbol{\omega}_i}L_i(\boldsymbol{\mu},\boldsymbol{\omega}) f(\boldsymbol{\mu},\boldsymbol{\omega}_i,\boldsymbol{\omega}_o;\alpha_{\boldsymbol{\mu}},\gamma_{\boldsymbol{\mu}},\mathbf{n}_{\boldsymbol{\mu}})(\boldsymbol{\omega}_i\cdot \mathbf{n}_\mu)\Delta\boldsymbol{\omega}_i,
\end{split}
\label{eq: pbr discrete}
\end{equation}
where ${\boldsymbol{\mu}}$ is the position of a 3D Gaussian, $\alpha_{\boldsymbol{\mu}}$, $\gamma_{\boldsymbol{\mu}}$ and $\mathbf{n}_{\boldsymbol{\mu}}$ is the corresponding albedo, roughness and normal attribute.
Finally, the physically-based rendered image can be obtained by rasterization (Eq.~\ref{eq: rasterization}).

\subsection{Training}
The optimizable parameters of the avatar model include the parameters of StyleUNets and an environment light map $\boldsymbol{l}_\text{env}$.
Our training loss consists of five parts: the reconstruction loss $\mathcal{L}_\text{recon}$, PBR loss $\mathcal{L}_\text{PBR}$, normal loss $\mathcal{L}_\text{normal}$, visibility loss $\mathcal{L}_\text{vis}$ and regularization loss $\mathcal{L}_\text{reg}$:
\begin{equation}
    \mathcal{L} = \mathcal{L}_\text{recon} + \mathcal{L}_\text{PBR}+\lambda_\text{normal}\mathcal{L}_\text{normal}+\lambda_\text{vis}\mathcal{L}_\text{vis}+\mathcal{L}_\text{reg}.
\end{equation}

\subsubsection{Reconstruction Loss}
The reconstruction loss aims to learn the geometry and the observed texture of the human avatar from the multi-view videos without considering the PBR process.
We denote the rendered image using color attributes from the Gaussian maps as $\mathbf{C}_\text{OBS}$.
The reconstruction loss involves an L1 loss and a perceptual loss \cite{zhang2018unreasonable} between $\mathbf{C}_\text{OBS}$ and the ground-truth image $\mathbf{C}_\text{GT}$:
\begin{equation}
    \mathcal{L}_\text{recon}=\mathcal{L}_1(\mathbf{C}_\text{OBS},\mathbf{C}_\text{GT}) + \lambda_\text{perceptual}\mathcal{L}_\text{perceptual}(\mathbf{C}_\text{OBS},\mathbf{C}_\text{GT}).
\end{equation}

\subsubsection{PBR Loss}
The PBR loss aims to disentangle the light condition and avatar materials from the multi-view observations.
We denote the rendered image by the PBR process (i.e., Eq.~\ref{eq: pbr discrete}) as $\mathbf{C}_\text{PBR}$.
The PBR loss involves an L1 loss and a perceptual loss between $\mathbf{C}_\text{PBR}$ and $\mathbf{C}_\text{GT}$:
\begin{equation}
    \mathcal{L}_\text{PBR}=\mathcal{L}_1(\mathbf{C}_\text{PBR},\mathbf{C}_\text{GT}) + \lambda_\text{perceptual}\mathcal{L}_\text{perceptual}(\mathbf{C}_\text{PBR},\mathbf{C}_\text{GT}).
\end{equation}

\subsubsection{Normal Loss}
We employ a pretrained normal estimation network \cite{saito2020pifuhd} to supervise our predicted normal.
Specifically, we render a normal image with the predicted normal as additional channels by rasterization, and compare it with the estimated one by \cite{saito2020pifuhd} using L1 loss.

\subsubsection{Visibility Loss}
In each iteration, for each 3D Gaussian, we randomly sample a light direction over the hemisphere around its normal.
We compute L1 loss between the predicted visibility of the sampled directions and the ground-truth one obtained by point-based ray tracing\cite{gao2023relightable}.

\subsubsection{Regularization Loss}
For the stability and convergence of avatar training, we design several regularization losses on the geometry and materials.
Specifically, the regularization losses involve a geometric regularization loss and smooth regularization losses on albedo and roughness:
\begin{equation}
    \mathcal{L}_\text{reg}=\lambda_\text{geo}\mathcal{L}_\text{geo}+\lambda_\text{albedo}\mathcal{L}_\text{smooth}(\mathbf{I}_\mathcal{A})+\lambda_\text{roughness}\mathcal{L}_\text{smooth}(\mathbf{I}_{\boldsymbol{\gamma}}),
\end{equation}
where $\mathcal{L}_\text{geo}=\|\Delta\mathcal{O}\|_2^2$ restrains the predicted offset map $\Delta\mathcal{O}$ from being extremely large, and the smooth loss is a bilateral smoothness term \cite{gao2023relightable} that constrains the material properties to change continuously in areas with smooth colors:
\begin{equation}
    \mathcal{L}_\text{smooth}(\mathbf{I})=\|\nabla \mathbf{I}\|\exp(-\|\nabla \mathbf{C}_\text{GT}\|),
\end{equation}
where $\mathbf{I}$ is the rendered albedo or roughness map ($\mathbf{I}_{\mathcal{A}}$ or $\mathbf{I}_{\boldsymbol{\gamma}}$) via splatting-based rasterization.

\subsection{Animation and Relighting}

\subsubsection{Pose Projection Strategy}
\label{subsec: pose projection}
Benefiting from the effective avatar representation, our method can reconstruct detailed human appearances under the training poses.
However, given the inherently data-driven nature of learning-based avatars, addressing generalization to novel poses is also necessary and important.
RAM-Avatar \cite{deng2024ramavatar} trains a VAE to transform the testing pose into an in-distribution one.
In this work, we propose to utilize Principal Component Analysis (PCA) to project a novel driving pose signal into the distribution of seen training poses for better generalization.
Specifically, given a pose condition represented by posed position maps, we extract valid points and concatenate them as a vector $\mathbf{x}_t\in\mathbb{R}^{3M}$ ($M$ is the point number). 
The vector of each training frame composes a matrix $\mathbf{X}=[\mathbf{x}_1,\cdots,\mathbf{x}_T]$, where $T$ is the number of training frames.
We perform PCA on $\mathbf{X}$, producing $N$ principal components $\mathbf{S}=[\mathbf{s}_1,\cdots,\mathbf{s}_N]\in\mathbb{R}^{3M\times N}$ and standard deviation of each component $\sigma_i$.
Given position maps derived by a novel driving pose, we project the corresponding vector $\mathbf{x}$ into the PCA space by
\begin{equation}
    \boldsymbol{\beta}=\mathbf{S}^\top \cdot (\mathbf{x}-\bar{\mathbf{x}}),
    \label{eq: pca proj}
\end{equation}
where $\bar{\mathbf{x}}$ is the mean of $\mathbf{X}$.
Then we reconstruct the positions from the low-dimensional coefficient $\boldsymbol{\beta}$ by
\begin{equation}
    \mathbf{x}_\text{recon}=\mathbf{S}\cdot \boldsymbol{\beta} + \bar{\mathbf{x}},
    \label{eq: pca recon}
\end{equation}
then we reshape $\mathbf{x}_\text{recon}$ into a $M\times 3$ tensor, and scatter it onto the position maps.
To constrain the reconstructed position maps to lie in the distribution of training poses, we clip each component of $\boldsymbol{\beta}$ within the bound of $[-2\sigma_i,2\sigma_i]$.
Overall, the pose projection strategy ensures reasonable interpolation within the distribution of training poses, enabling better generalization to novel poses as shown in Fig.~\ref{fig: eval pose projection}.

\subsubsection{Relighting}
Thanks to the introduction of PBR into the avatar representation, our method is able to relight the avatar under novel illumination.
Specifically, given a novel pose, we first feed it into the avatar networks to predict Gaussian, normal, albedo and roughness maps.
Note that we do not predict light visibility maps during testing because we empirically find it cannot accurately generalize to novel poses.
We hypothesize the reasons are that the visibility is heavily determined by global self-occlusions, and the variety of training poses is limited.
To this end, we directly calculate the visibility of posed 3D Gaussians via point-based ray tracing \cite{gao2023relightable}.
Then, given a novel environment map, we inject it with the predicted normal, abledo, and roughness as well as calculated visibility into the PBR equation (Eq.~\ref{eq: pbr discrete}) to compute the PBR color of each 3D Gaussian.
Finally, the relit image is obtained by splatting-based rasterization.
As illustrated in Fig.~\ref{fig: relit results}, our method can produce photorealistic animation under novel illumination.

%% file: secs/4_results.tex
\begin{figure*}[t]
    \centering
    \includegraphics[width=\linewidth]{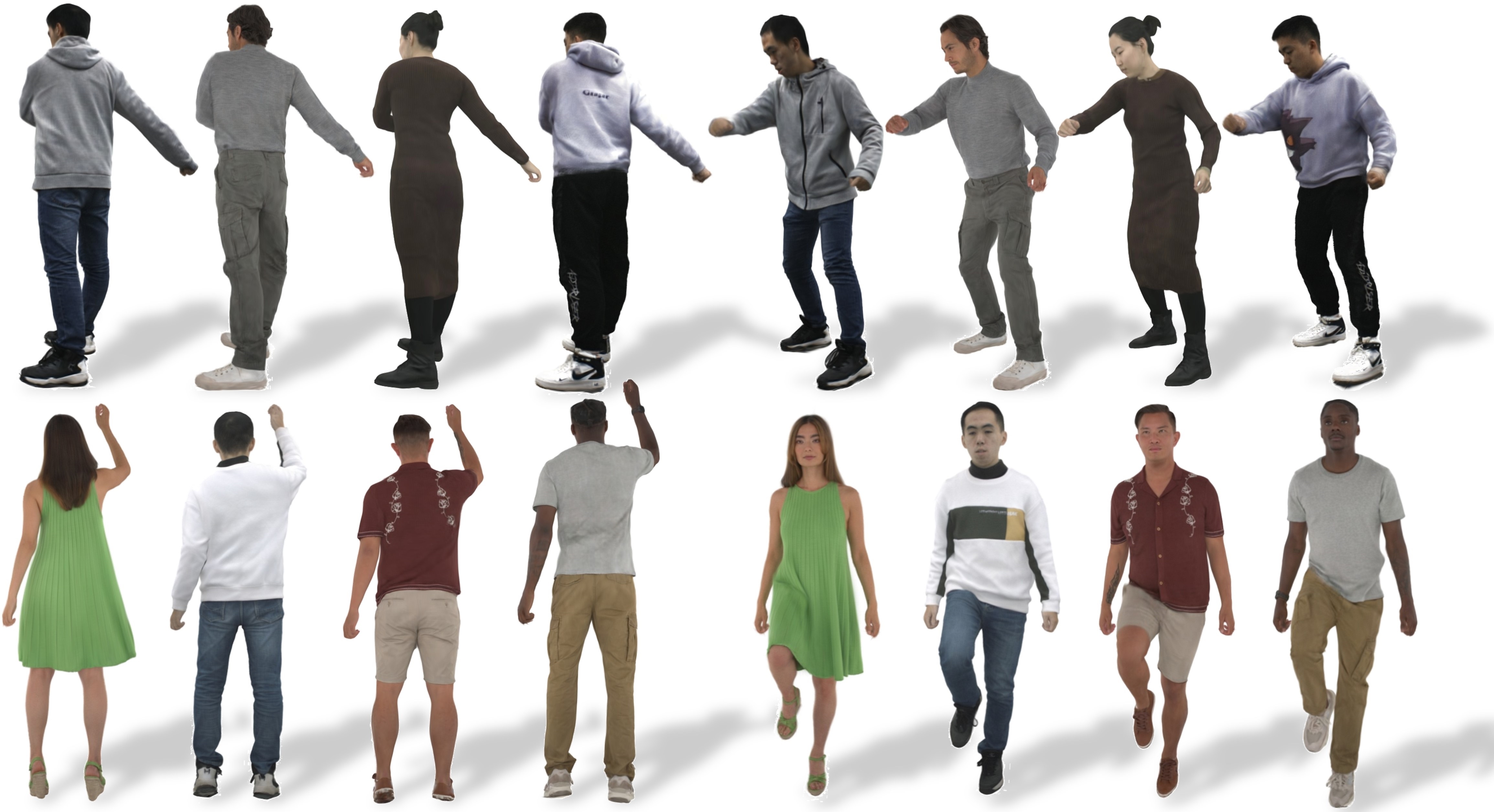}
    \caption{\textbf{Example animatable avatars with high-fidelity dynamic appearances created by our method.}}
    \label{fig: results}
\end{figure*}

\section{Experiments}

\subsection{Results}

\subsubsection{Animation}
\begin{figure*}[htbp]
    \centering
    \includegraphics[width=\linewidth]{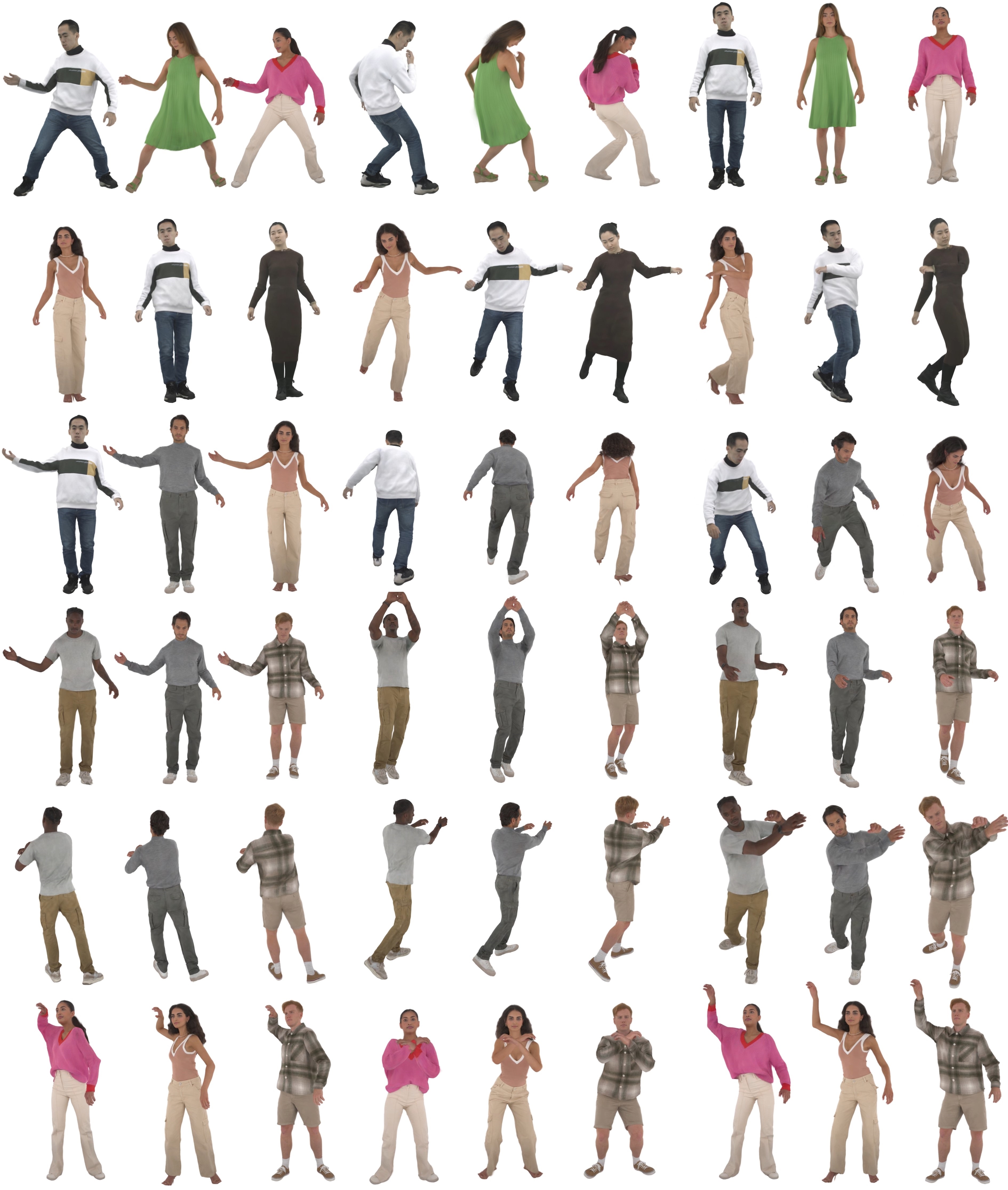}
    \caption{\textbf{Example sequential animation results by our method.}
    Each row is an animation sequence involving 3 subjects.
    Our method can generate realistic and reasonable dynamic details even under novel poses from the AMASS dataset \cite{mahmood2019amass}.}
    \label{fig: amass results}
\end{figure*}

As shown in Fig.~\ref{fig: teaser} and Fig.~\ref{fig: results}, our method can create realistic avatars with high-fidelity dynamic details from multi-view videos.
We also show results animated by challenging out-of-distribution poses from AMASS dataset \cite{mahmood2019amass} in Fig.~\ref{fig: amass results}.
Thanks to the effective avatar representation and pose projection strategy, our method can produce animation results with highly dynamic, realistic and generalized appearances.

\begin{figure*}[!t]
    \centering
    \includegraphics[width=\linewidth]{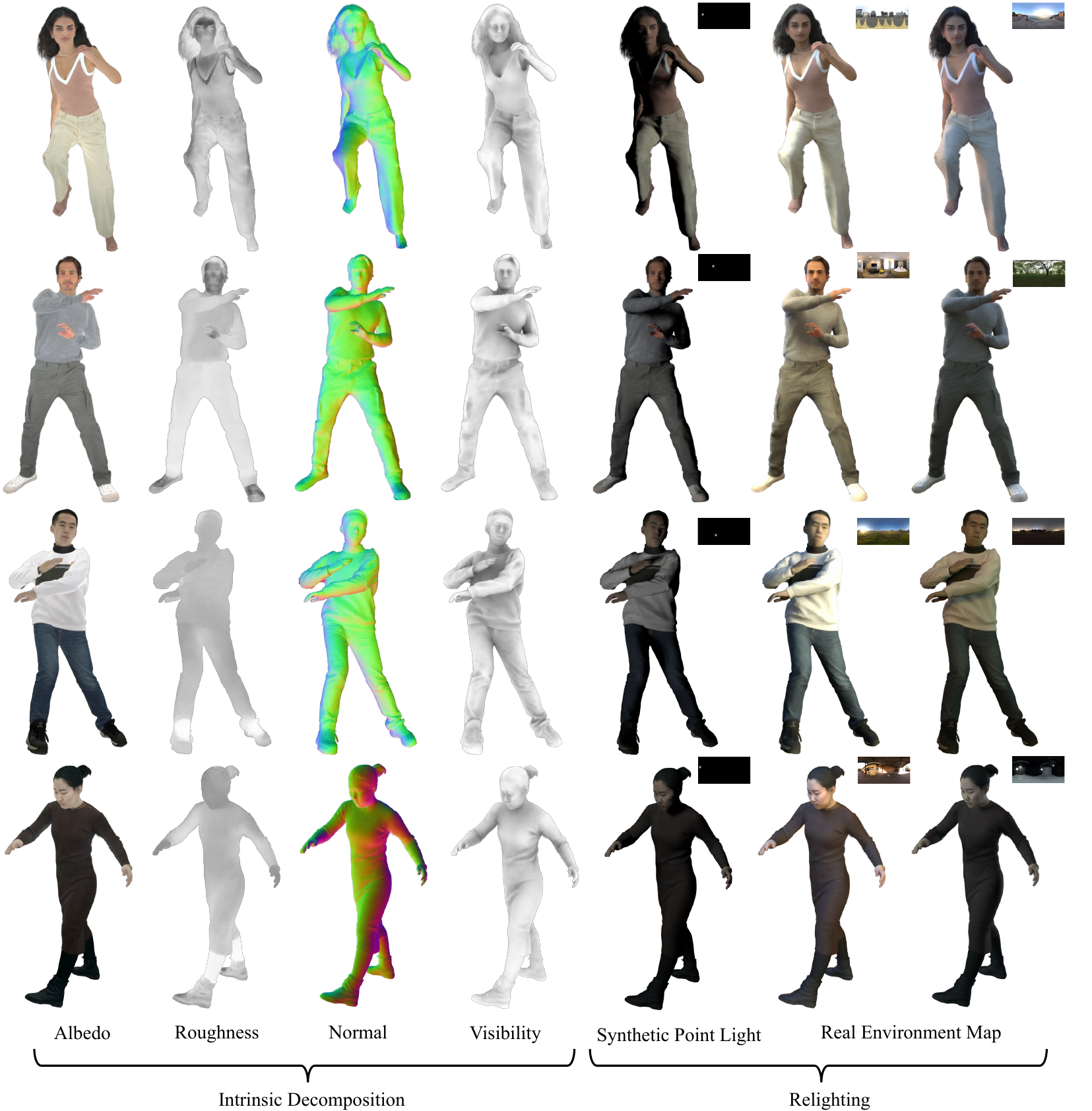}
    \caption{\textbf{Example intrinsic decomposition and relighting results under novel poses by our method.}
    The first four columns display the decomposition results of material and geometry. The fifth column showcases the relighting outcomes with a directional point light, whereas the remaining columns are relit using real environment maps.}
    \label{fig: relit results}
\end{figure*}

\subsubsection{Relighting}
In addition to animation, our method extends support to relighting applications. By naturally incorporating physically-based inverse rendering techniques into our powerful avatar representation, we achieve accurate intrinsic decomposition and vivid relighting results, as demonstrated in Fig.~\ref{fig: teaser} and Fig.~\ref{fig: relit results}.

Please refer to the supplementary video for more sequential animation and relighting results.

\subsection{Dataset and Metric}
We mainly utilize three public datasets for the experiments, including 3 sequences with 24 views from THuman4.0 dataset \cite{zheng2022structured}, 3 sequences with 16 views from AvatarReX dataset \cite{zheng2023avatarrex} and 5 sequences with 160 views from ActorsHQ dataset \cite{isik2023humanrf} (we only use 47 full-body views for avatar modeling).
THuman4.0 and AvatarReX datasets also provide the SMPL-X \cite{SMPL-X:2019} registrations.
We fit SMPL-X for ActorsHQ dataset using the method proposed by Zhang \etal \cite{zhang2021lightweight}.
We split each sequence as training and testing chunks, and the training chunk contains 1500 $\sim$ 3000 frames.

We adopt Peak Signal-to-Noise Ratio (PSNR), Structure
Similarity Index Measure (SSIM) \cite{wang2004image}, Learned Perceptual Image Patch Similarity (LPIPS) \cite{zhang2018unreasonable} and
Fréchet Inception Distance (FID) \cite{heusel2017gans} for
quantitative experiments.
PSNR and SSIM are computed on the entire image at the original resolution, while LPIPS and FID are computed on the cropped minimal square that covers the human body.

\begin{figure*}[t]
    \centering
    \includegraphics[width=\linewidth]{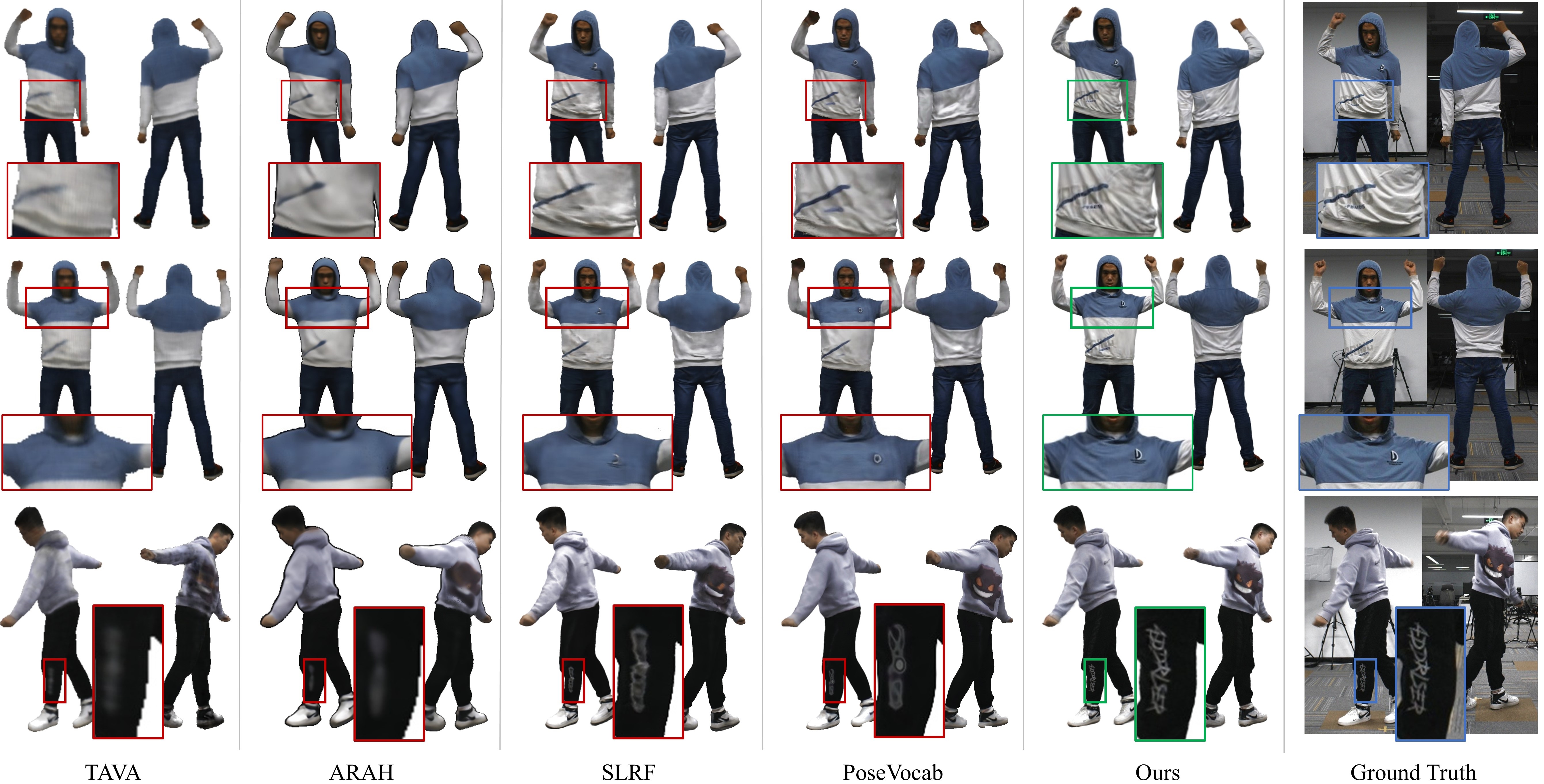}
    \caption{\textbf{Qualitative comparison with state-of-the-art NeRF-based body-only avatars including TAVA \cite{li2022tava}, ARAH \cite{wang2022arah}, SLRF \cite{zheng2022structured} and PoseVocab \cite{li2023posevocab} on novel pose synthesis.}}
    \label{fig: comparison body-only}
\end{figure*}

\begin{table}[t]
\caption{\textbf{Loss weights in the training process.}}
\label{tab: loss weights}
\centering
\begin{tabular}{lc}
\hline
Loss Weight                & Value \\ \hline
$\lambda_\text{perceptual}$ & 0.1 \\
$\lambda_\text{normal}$     & 0.2 \\
$\lambda_\text{vis}$        & 0.1 \\
$\lambda_\text{geo}$        & 0.01 \\
$\lambda_\text{albedo}$     & 0.005 \\
$\lambda_\text{roughness}$  & 0.005 \\ \hline
\end{tabular}
\end{table}

\subsection{Implementation Details}

\subsubsection{Template Reconstruction}
We optimize an SDF and color field represented by an MLP consisting of intermediate layers with (512, 256, 256, 256, 256, 256) neurons.
Given a posed point, we find accurate correspondence in the canonical space by root finding.
Following ARAH \cite{wang2022arah}, we initialize the correspondence as the canonical position that is computed by inverse skinning based on blending weights of the closest SMPL vertex.
Different from SNARF \cite{chen2021snarf} and ARAH \cite{wang2022arah} that utilize the Broyden's method \cite{broyden1965class} to solve Eq.~\ref{eq: root finding}, we employ the Gauss-Newton method by implementing a customized CUDA kernel.
The training loss of template reconstruction involves an RGB loss, a mask loss and an Eikonal loss \cite{gropp2020implicit}.

\subsubsection{Network Architecture}
The network in our avatar representation is composed of StyleUNet \cite{wang2023styleavatar}, a conditional StyleGAN-based \cite{karras2019style} generator.
Differently, we adapt the original StyleUNet by incorporating two decoders to predict both front \& back Gaussian and intrinsic maps.
The resolution of the input position map is $512\times 512$, and the resolution of the output Gaussian and intrinsic maps is $1024 \times 1024$.
Specifically, we utilize five different StyleUNets to output color (3-channel), position (3-channel), other Gaussian attributes (8-channel), albedo \& roughness (4-channel) and normal \& visibility (19-channel).
The visibility is represented as 16-dimensional SH coefficients.
In the color StyleUNet, we modulate the color output with a view direction map to model view-dependent effects.
Each pixel on the view direction map indicates the angle between the view direction and the template normal.
The view direction map is encoded through a tiny CNN, then the encoded feature map is injected into an intermediate decoder layer of the color StyleUNet.

\subsubsection{Training}
We adopt the Adam optimizer \cite{adam} for training the network with a learning rate of $5\times 10^{-4}$.
The loss weights are set as illustrated in Tab.~\ref{tab: loss weights}.
The batch size is 1, the total iteration number is 500k, and the training procedure takes about two days on one RTX 4090.

\subsubsection{Running Time}
For only animation, it takes around 0.13 secs to render one frame.
Given novel illumination for relighting, it takes $4\sim 10$ secs to synthesize one frame due to the additional computational cost of light visibility computation in PBR.
The relighting time cost mainly depends on the numbers of 3D Gaussians and sampled rays.

\input{tabs/comparison_body_only}

\subsection{Comparison on Avatar Animation}
In the comparisons with state-of-the-art animatable avatars, we first compare our method with NeRF-based approaches including both body-only (TAVA \cite{li2022tava}, ARAH \cite{wang2022arah}, SLRF \cite{zheng2022structured}, PoseVocab \cite{li2023posevocab}) and full-body (AvatarReX \cite{zheng2023avatarrex}) avatars.
Then we compare our method with concurrent 3D Gaussian splatting-based avatars including 3DGS-Avatar \cite{qian20233dgs} and GaussianAvatar \cite{hu2023gaussianavatar}.

\subsubsection{NeRF-based Body-only Avatars}
We compare our method with TAVA, ARAH, SLRF, and PoseVocab on ``subject00'' and ``subject02'' sequences of THuman4.0 dataset \cite{zheng2022structured}.
We run the released codes of TAVA, ARAH and PoseVocab on the dataset, and request the results of SLRF from the authors.
We present qualitative comparisons on novel pose synthesis in Fig.~\ref{fig: comparison body-only}. 
In contrast to other methods, our approach excels in animating highly realistic avatars with significant improvement on high-fidelity dynamic details, including garment wrinkles, logos and other textural patterns.
The quantitative comparison is also performed on the testing chunk (the 2000-2500 frames and ``cam18'' view) of the ``subject00'' sequence as shown in Tab.~\ref{tab: comparison body-only}, and these numerical results prove that our method achieves more accurate animation.
Although PoseVocab and SLRF introduce a learnable pose dictionary or local NeRFs to improve the representation ability of the NeRF MLP, they still suffer from the low-frequency bias \cite{tancik2020fourier} of MLPs and fail to create highly realistic avatars.
Contrarily, our method leverages powerful 2D CNNs and explicit 3D Gaussian splatting, thus achieving modeling finer-grained dynamic appearances.

\begin{figure}[t]
    \centering
    \includegraphics[width=\linewidth]{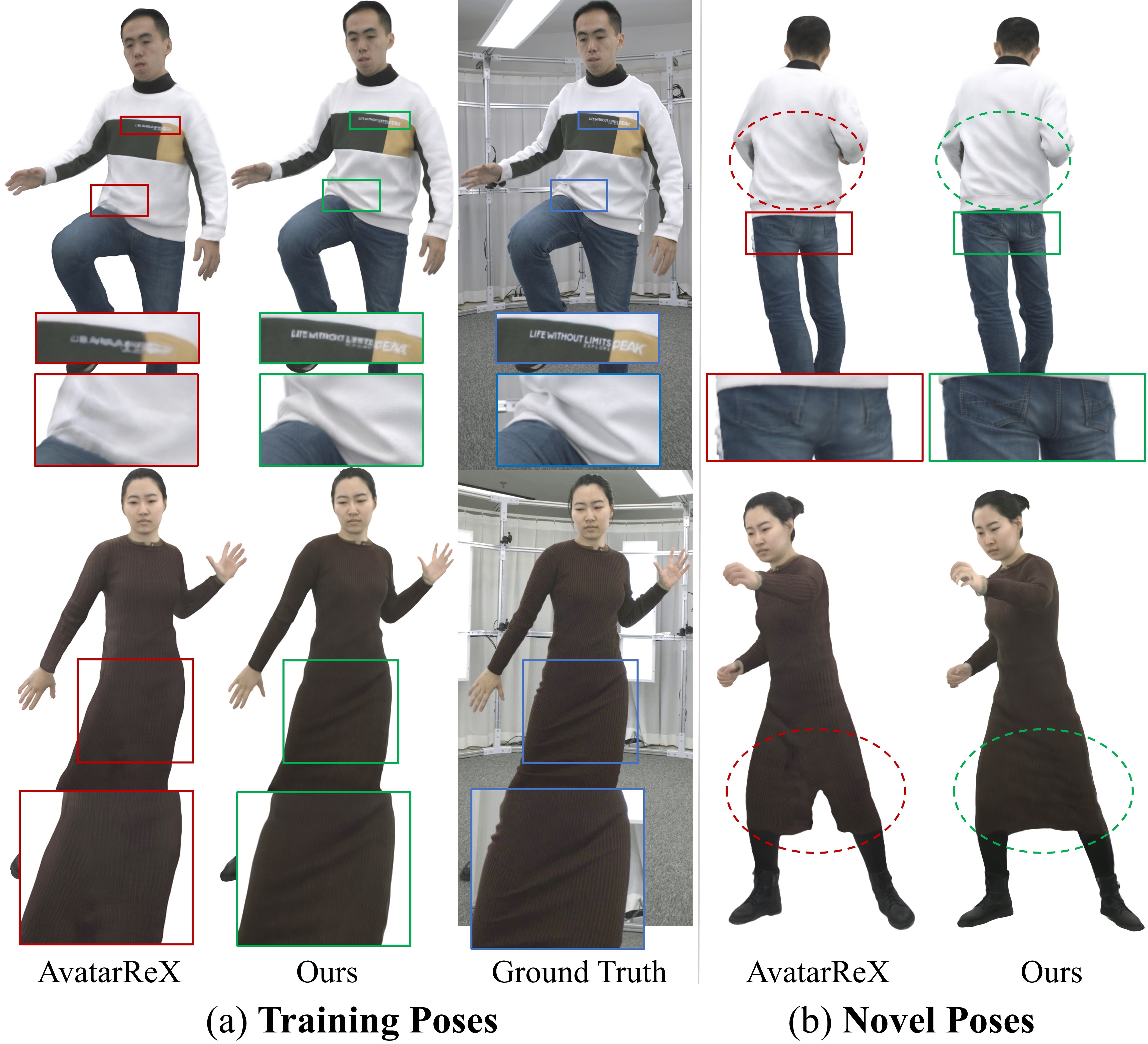}
    \caption{\textbf{Qualitative comparison with AvatarReX \cite{zheng2023avatarrex} on both training pose reconstruction (a) and novel view synthesis (b).}}
    \label{fig: comparison full-body}
\end{figure}

\input{tabs/comparison_full_body}

\subsubsection{NeRF-based Full-body Avatars}
Full-body avatars including TotalSelfScan \cite{dong2022totalselfscan}, X-Avatar \cite{shen2023xavatar} and AvatarReX \cite{zheng2023avatarrex} can realize expressive control of the body, hands and face.
TotalSelfScan reconstructs full-body avatars from monocular self-rotation videos, and only displays animations that appear very rigid.
X-Avatar requires 3D human scans under different poses as input for creating avatars.
AvatarReX is the most relevant work with our method, i.e., creating avatars from multi-view videos.
Fig.~\ref{fig: comparison full-body} shows the comparison with AvatarReX on both training and novel poses.
Fig.~\ref{fig: comparison full-body} (a) demonstrates that our method can reconstruct more faithful and vivid details compared with AvatarReX.
Although AvatarReX introduces local feature patches to encode more details, it remains constrained by the representation ability of the conditional NeRF MLPs.
Fig.~\ref{fig: comparison full-body} (b) shows that given a novel pose, our method not only generates more realistic details but also produces more reasonable non-rigid deformation, particularly for long dresses, in comparison with AvatarReX.
This is attributed to the ability of our method to learn pose-dependent deformations on a character-specific template that has already modeled the basic shape of the wearing garments. In contrast, AvatarReX learns sparse node translations on the naked SMPL model, resulting in artifacts for long dresses.
Tab.~\ref{tab: comparison full-body} reports the quantitative comparison on training pose reconstruction. Our method also outperforms AvatarReX on the reconstruction accuracy.
The numerical results are evaluated on the first 500 frames and the ``22010710'' camera view in the ``avatarrex\_zzr'' sequence from AvatarReX dataset.

\begin{figure}[t]
    \centering
    \includegraphics[width=\linewidth]{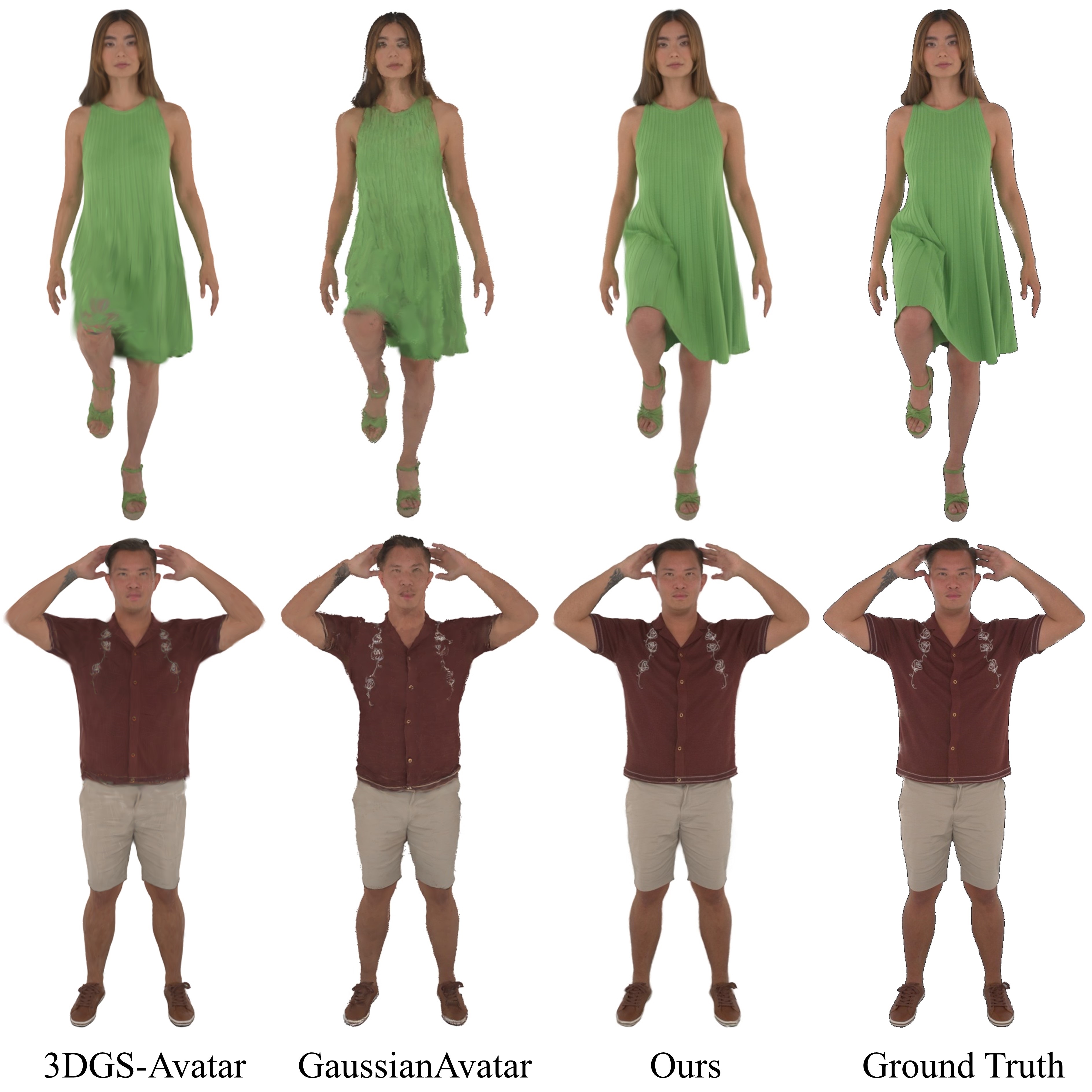}
    \caption{\textbf{Qualitative comparison with state-of-the-art 3D Gaussian splatting-based avatars including 3DGS-Avatar \cite{qian20233dgs} and GaussianAvatar \cite{hu2023gaussianavatar}.}}
    \label{fig: comparison 3dgs avatars}
\end{figure}

\input{tabs/comparison_3dgs_avatars}

\begin{figure*}[t!]
    \centering
    \includegraphics[width=\linewidth]{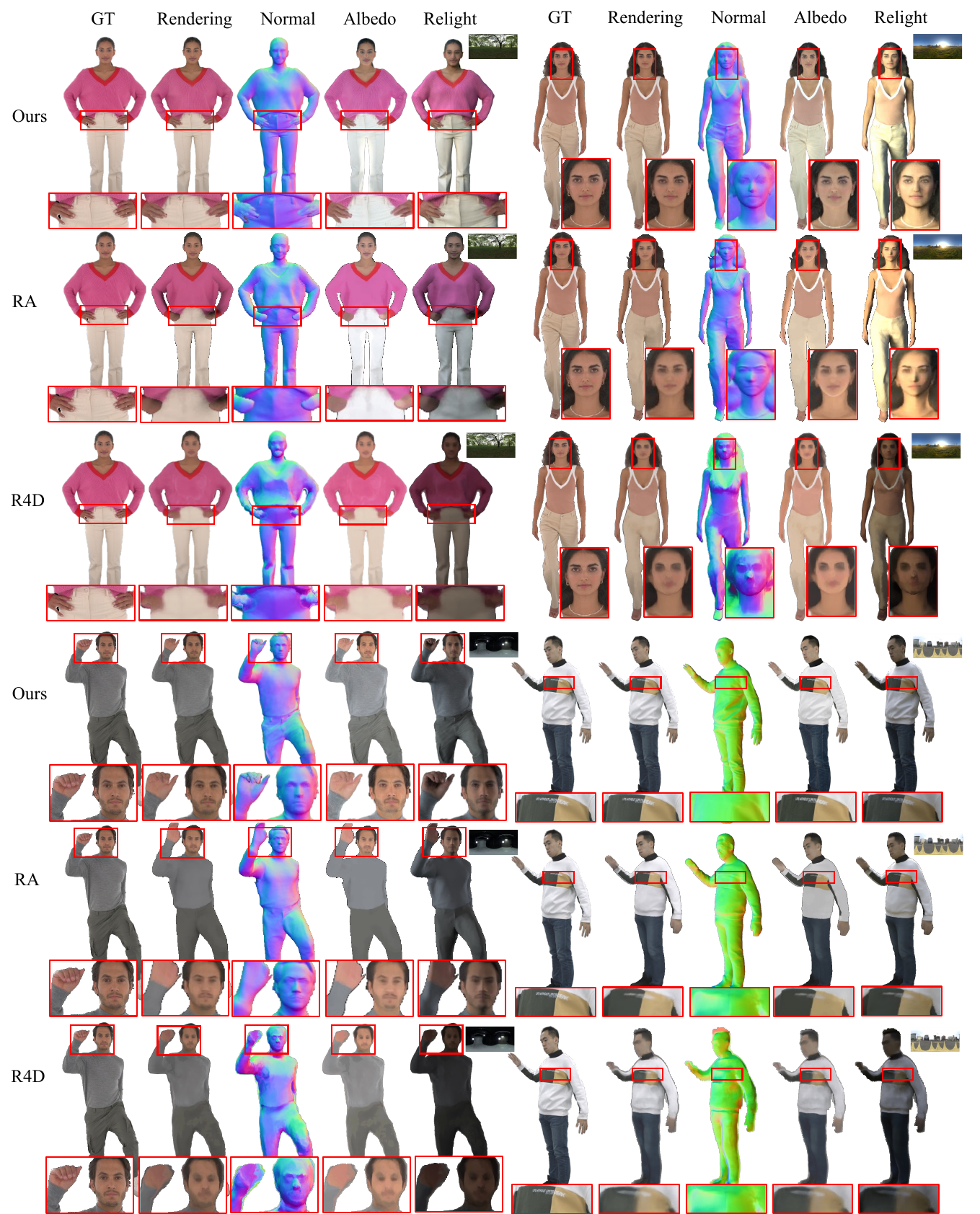}
    \caption{\textbf{Qualitative comparison with the state-of-the-art human performance relighting methods, R4D \cite{chen2022relighting4d} and RA \cite{lin2024relightable}.}}
    \label{fig: comparison relit}
\end{figure*}

\subsubsection{3D Gaussian Splatting-based Avatars}
We qualitatively compare our method with 3DGS-Avatar \cite{qian20233dgs} and GaussianAvatar \cite{hu2023gaussianavatar} using ``Actor01'' and ``Actor02'' sequences from ActorsHQ dataset \cite{isik2023humanrf} in Fig.~\ref{fig: comparison 3dgs avatars}.
It shows that our method outperforms other approaches by a large margin on the avatar quality, especially the dynamic wrinkles of the garments.
3DGS-Avatar \cite{qian20233dgs} utilizes pure MLPs to regress the non-rigid deformations and pose-dependent appearances, suffering from the limited capacity of MLPs.
Although GaussianAvatar \cite{hu2023gaussianavatar} employs a 2D U-Net \cite{ronneberger2015u} to regress Gaussian parameters on SMPL UV space, it freezes the opacity and rotation attribute as pose-agnostic variables, deteriorating the modeling ability of the whole model.
Moreover, these two methods utilize the naked SMPL model to parameterize 3D Gaussians and both fail to model detailed motions and appearances of loose clothes as shown in the top row of Fig.~\ref{fig: comparison 3dgs avatars}.
We also report the numerical results in Tab.~\ref{tab: comparison 3dgs avatars}, and our method also quantitatively outperforms 3DGS-Avatar and GaussianAvatar.
The numerical results are computed on the 48-548 frames and the ``Cam127'' camera view in the ``Actor01/Sequence1'' from ActorsHQ dataset.

\input{tabs/comparison_relit}

\subsection{Comparison on Human Performance Relighting}
We compare our method with the state-of-the-art human performance relighting methods Relighting4D (R4D) \cite{chen2022relighting4d} and Relightable and Animatable Avatar (RA) \cite{lin2024relightable} on ActorsHQ \cite{isik2023humanrf} and AvatarReX \cite{zheng2023avatarrex} datasets. 

We present the qualitative and quantitative comparison in Fig.~\ref{fig: comparison relit} and Tab.~\ref{tab: comparison relit}. Our method excels in precise intrinsic decomposition due to the proposed physically-based avatar representation, particularly evident in the albedo and normal details. This highlights the enhanced capabilities of our approach in capturing intricate details in both geometry and appearance. Moreover, the accurate material and geometry can then contribute to natural relighting under novel illumination. However, R4D \cite{chen2022relighting4d} and RA \cite{lin2024relightable}, as NeRF-based implicit methods, fail to model detailed material and geometry of the character due the limited capacity of their avatar representations. 
These two approaches produce blurry results on both texture and geometry, leading to diminished relighting results.

\subsection{Ablation Study}
We evaluate the core contributions of our method in this subsection.

\begin{figure}[t!]
    \centering
    \includegraphics[width=\linewidth]{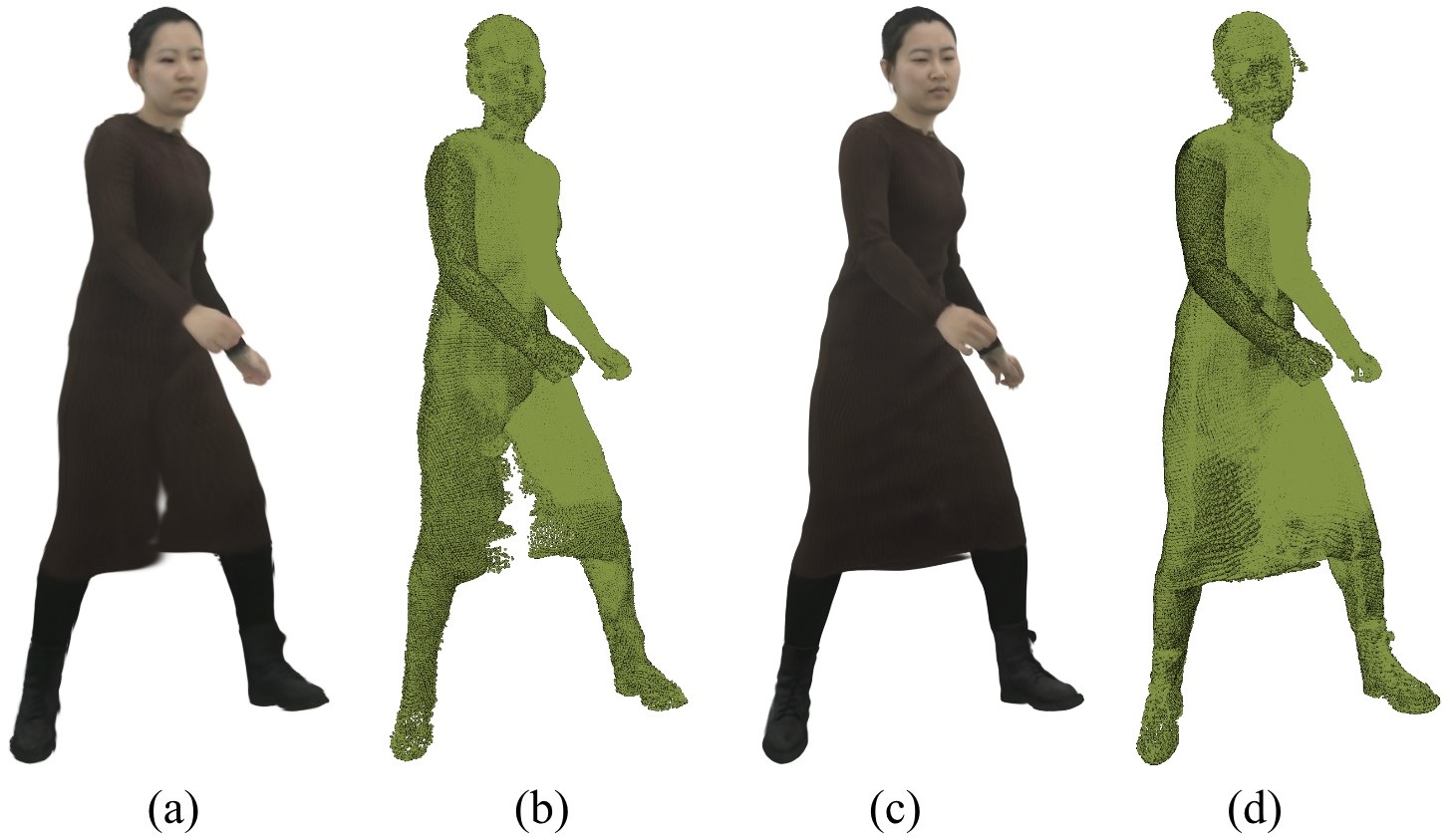}
    \caption{\textbf{Ablation study of the parametric template.}
    (a,b) Rendered results and 3D Gaussians using SMPL-X. (c,d) Rendered results and 3D Gaussians using the character-specific template.
    }
    \label{fig: eval template}
\end{figure}

\subsubsection{Parametric Template}
We evaluate the learned parametric template by replacing it with a naked parametric model, SMPL-X \cite{SMPL-X:2019}.
Fig.~\ref{fig: eval template} shows that SMPL-X fails to represent the long dress whose topology is not consistent with the SMPL-X model, yielding poor generalization to novel poses.
Conversely, our character-specific template is adaptively reconstructed from the input video to model the basic shape of the wearing garments.
We also quantitatively compare the reconstructed parametric template with the naked SMPL-X model \cite{SMPL-X:2019} on the animation accuracy in Tab.~\ref{tab: quant eval template}.
It shows that the reconstructed template can animate the 3D Gaussians more accurately.

\begin{table}[t!]
\caption{\textbf{Quantitative ablation study on the parametric template.}
}
\label{tab: quant eval template}
\centering
\begin{tabular}{lcccc}
\hline
                 & PSNR $\uparrow$   & SSIM $\uparrow$  & LPIPS $\downarrow$ & FID $\downarrow$  \\ \hline
Parametric Template & \textbf{31.2183} & \textbf{0.9858} & \textbf{0.0344}   & \textbf{36.9905} \\
SMPL-X           & 30.5241          & 0.9842          & 0.0401            & 47.5066          \\ \hline
\end{tabular}
\end{table}

\begin{figure}[t!]
    \centering
    \includegraphics[width=\linewidth]{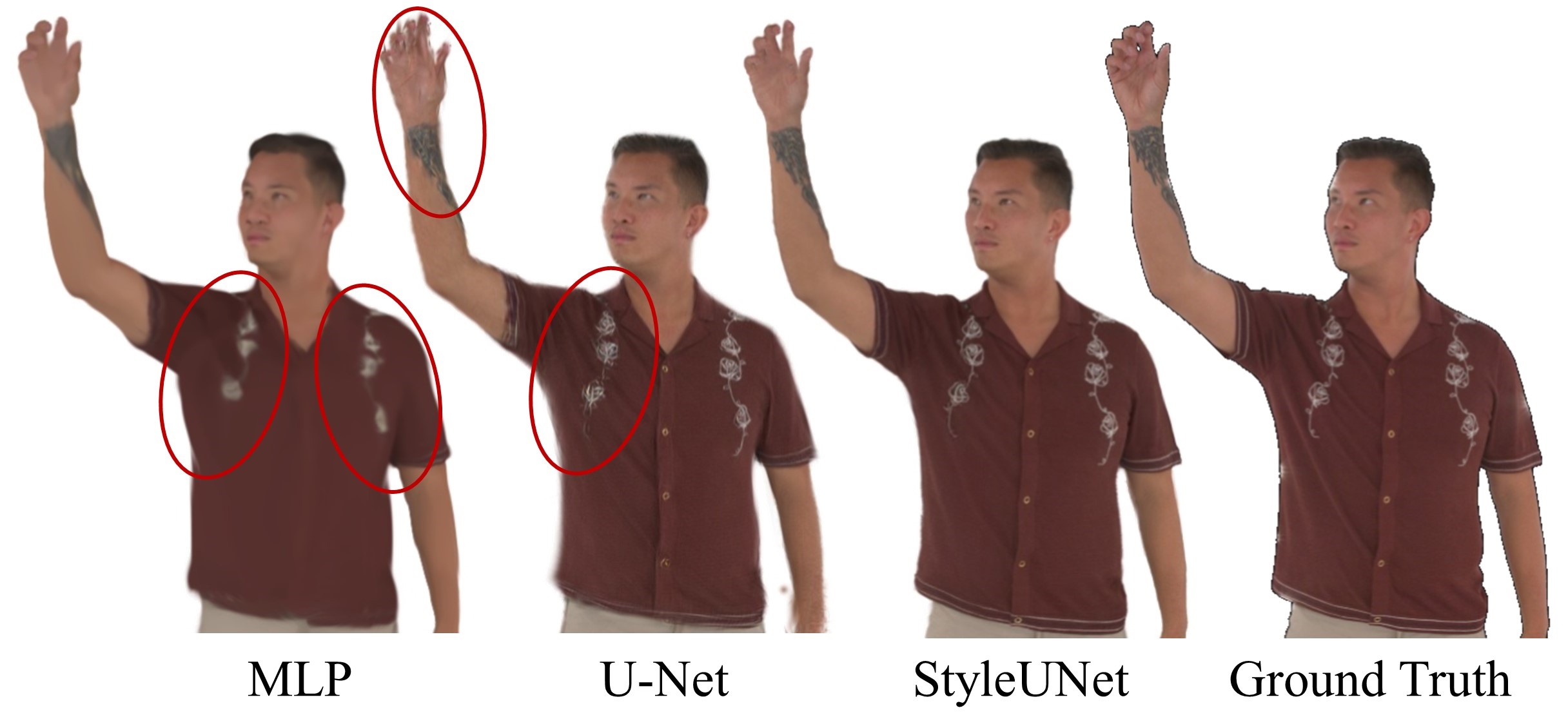}
    \caption{\textbf{Comparison between representations with different backbones on training pose reconstruction.}
    }
    \label{fig: eval backbones}
\end{figure}

\begin{table}[t!]
\centering
\caption{\textbf{Quantitative comparison between representations with different backbones.}}
\label{tab: quant eval backbones}
\footnotesize
\begin{tabular}{lcccc}
\hline
        & PSNR $\uparrow$   & SSIM $\uparrow$  & LPIPS $\downarrow$ & FID $\downarrow$  \\ \hline
StyleUNet \cite{wang2023styleavatar} & \textbf{29.3127} & \textbf{0.9664} & \textbf{0.0378}   & \textbf{27.3143} \\
U-Net \cite{ronneberger2015u} & 26.4255 & 0.9435 & 0.0507 & 31.3838\\
MLP    & 26.8961          & 0.9497          & 0.0650            & 87.0793          \\ \hline
\end{tabular}
\end{table}

\subsubsection{Backbones}
To demonstrate the superior representation ability of 2D CNNs (StyleUNet in our settings), we replace StyleUNet with a coordinate-based MLP and a standard U-Net \cite{ronneberger2015u}, respectively.
The MLP takes a canonical point and pose vector as input, and returns the 3D Gaussian attributes of this point.
While the standard U-Net replaces the StyleUNet as the backbone.
Fig.~\ref{fig: eval backbones} and Tab.~\ref{tab: quant eval backbones} show the qualitative and quantitative animation results of our method with StyleUNet and the baselines with MLPs and U-Net, respectively.
First, it demonstrates that 2D CNNs are able to regress more detailed and realistic appearances, while MLPs suffer from limited representation ability, yielding blurry animation results.
Second, StyleUNet outperforms the standard U-Net because of the additional modules (including style modulation and ``To/From-RGB'' modules) inherited from StyleGAN \cite{karras2019style}.
Overall, the 2D parameterization and StyleUNet enable our method to model high-quality human dynamic appearances.

\begin{figure}[t!]
    \centering
    \includegraphics[width=\linewidth]{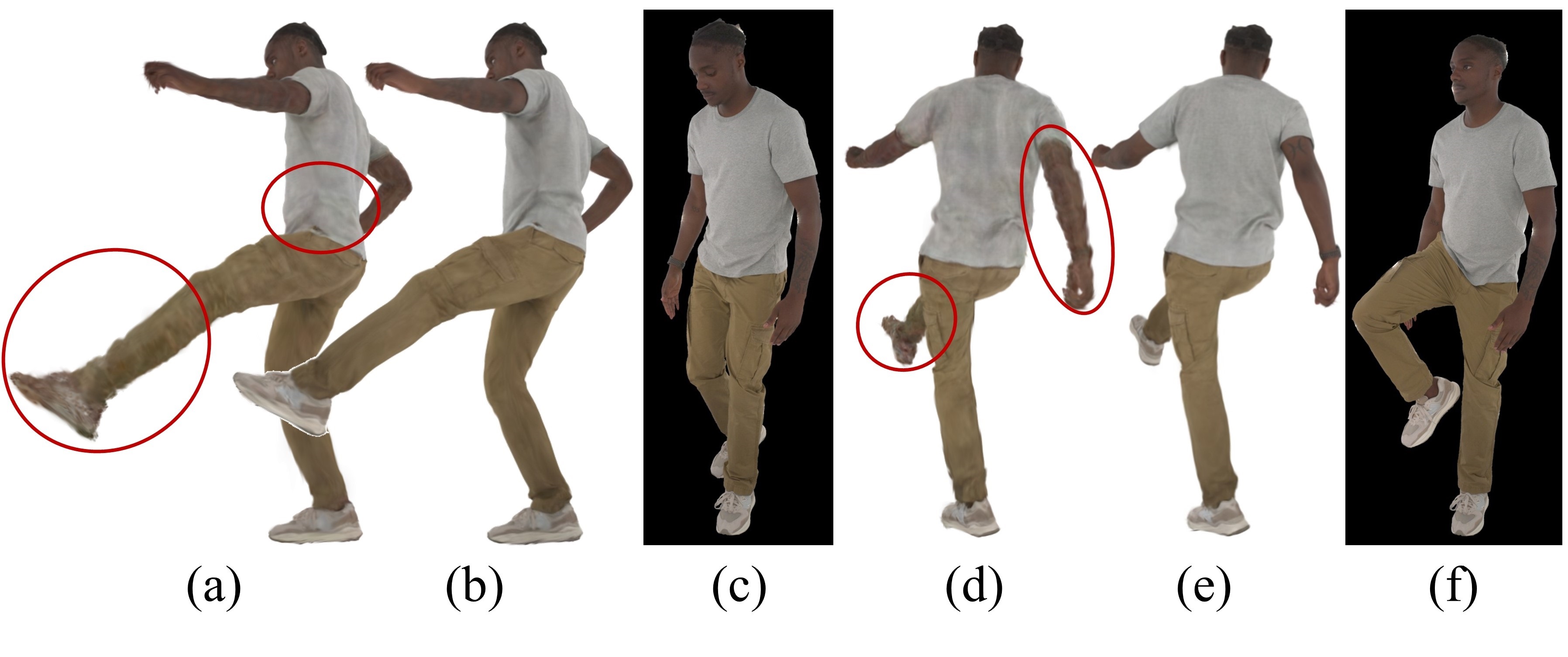}
    \caption{\textbf{Ablation study of the pose projection strategy.}
    (a,d) and (b,e) are the animation results without and with the pose projection strategy, respectively. (c,f) are the reference images with the closest pose in the training dataset.
    }
    \label{fig: eval pose projection}
\end{figure}

\begin{table}[t!]
\centering
\caption{\textbf{Quantitative ablation study on pose projection.}}
\label{tab: eval pose projection}
\begin{tabular}{lcccc}
\hline
               & PSNR $\uparrow$   & SSIM $\uparrow$  & LPIPS $\downarrow$ & FID $\downarrow$  \\ \hline
w Pose Proj.   & \textbf{24.9932} & \textbf{0.9285} & \textbf{0.0685}   & \textbf{45.6266} \\
w/o Pose Proj. & 23.5594          & 0.9189          & 0.0792            & 59.9083          \\ \hline
\end{tabular}
\end{table}

\begin{figure}[t!]
    \centering
    \includegraphics[width=0.95\linewidth]{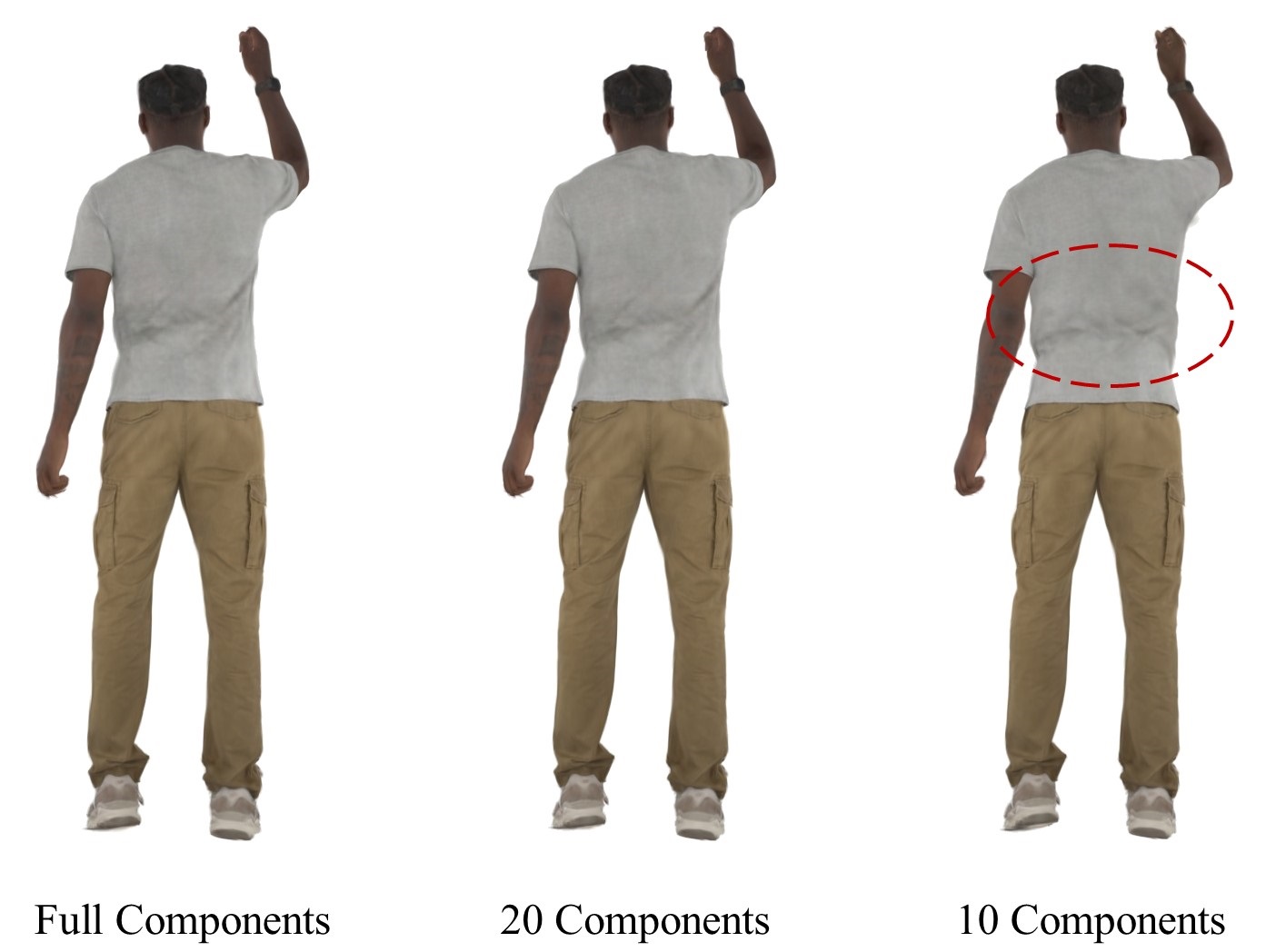}
    \caption{\textbf{Ablation study on the component number in the pose projection.}}
    \label{fig: supp n pca}
\end{figure}

\subsubsection{Pose Projection}
We evaluate the pose projection strategy by removing it, i.e., directly inputting the position map into the StyleUNet.
Fig.~\ref{fig: eval pose projection} and Tab.~\ref{tab: eval pose projection} show the qualitative and quantitative animation results with and without the pose projection under novel poses, respectively.
It demonstrates that direct extrapolation with the novel position map results in unreasonable 3D Gaussians, since no similar poses in the training dataset.
In contrast, the pose projection guarantees that the reconstructed position maps (Eq.~\ref{eq: pca recon}) lie within the distribution of training poses, leading to reasonable and vivid synthesized appearances.

\subsubsection{Number of Principal Components in Pose Projection}
Fig.~\ref{fig: supp n pca} shows the animation results with different numbers of principal components in the pose projection strategy.
It demonstrates that although PCA can project a novel pose into the distribution of the training poses for better pose generalization as shown in Fig.~\ref{fig: eval pose projection}, too few principal components may lose some fine-grained garment details.
We empirically found that setting the number of principal components to 20 could produce both detailed and generalized animation.

\subsubsection{View Number}
\begin{figure}[t]
    \centering
    \includegraphics[width=\linewidth]{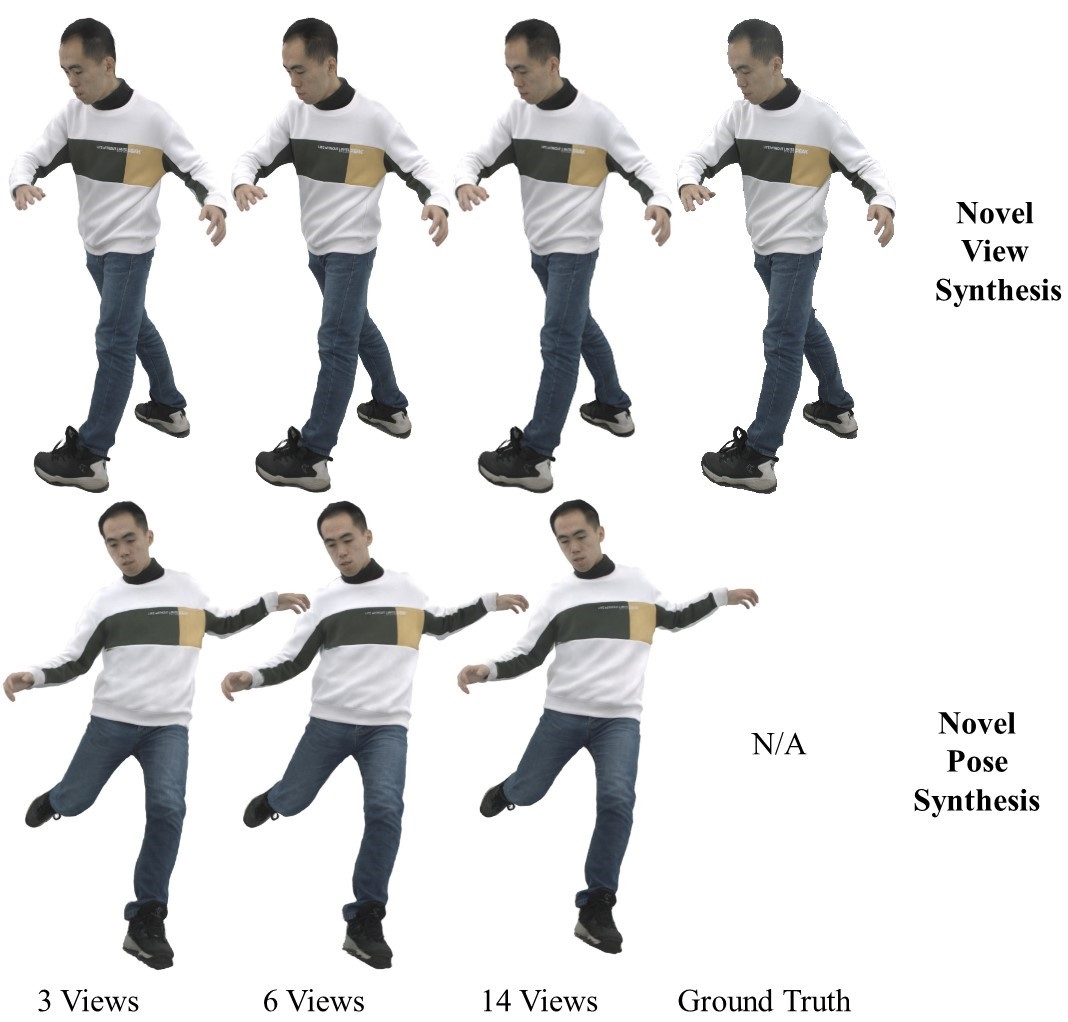}
    \caption{\textbf{Animation results trained with different numbers of views.}}
    \label{fig: eval view num}
\end{figure}

We quantitatively and qualitatively show the animation results trained with 3 views, 6 views and 14 views in Tab.~\ref{tab: eval view num} and Fig.~\ref{fig: eval view num}.
They demonstrate that our method also supports sparse-view input and can realize comparable high-fidelity results.

\input{tabs/ablation_sparse_view}

%% file: tabs/comparison_body_only.tex
\begin{table}[t]
\caption{\textbf{Quantitative comparison with state-of-the-art NeRF-based body-only avatars.}}
\label{tab: comparison body-only}
\centering
\small
\begin{tabular}{lllll}
\hline
Method    & \multicolumn{1}{c}{PSNR $\uparrow$} & SSIM $\uparrow$ & \multicolumn{1}{c}{LPIPS $\downarrow$} & \multicolumn{1}{c}{FID $\downarrow$} \\ \hline
Ours      & \textbf{28.0714}         & \textbf{0.9739} &  \textbf{0.0515}                   & \textbf{29.4831}                   \\
PoseVocab \cite{li2023posevocab} & 26.3784                             & 0.9707          & 0.0592                                 & 49.4541                              \\
SLRF \cite{zheng2022structured}     & 26.9015                             & 0.9724          & 0.0600                                 & 52.0613                            \\
ARAH \cite{wang2022arah}     & 22.3004                             & 0.9616          & 0.1075                                 & 90.6077                            \\
TAVA \cite{li2022tava}     & 26.8019                             & 0.9705          & 0.0915                                 & 96.3474                             \\ \hline
\end{tabular}
\end{table}

%% file: tabs/comparison_full_body.tex
\begin{table}[t]
\caption{\textbf{Quantitative comparison with the state-of-the-art NeRF-based full-body avatar.}}
\label{tab: comparison full-body}
\centering
\small
\begin{tabular}{lllll}
\hline
Method    & \multicolumn{1}{c}{PSNR $\uparrow$} & SSIM $\uparrow$ & \multicolumn{1}{c}{LPIPS $\downarrow$} & \multicolumn{1}{c}{FID $\downarrow$} \\ \hline
Ours      & \textbf{30.6143}                    & \textbf{0.9803} & \textbf{0.0290}                        & \textbf{13.2417}                     \\
AvatarReX \cite{zheng2023avatarrex} & 23.2475                             & 0.9567          & 0.0646                                 & 31.1387                             \\ \hline
\end{tabular}
\end{table}

%% file: tabs/comparison_3dgs_avatars.tex
 \begin{table}[t]
 \caption{\textbf{Quantitative comparison with state-of-the-art 3D Gaussian splatting-based avatars.}}
 \label{tab: comparison 3dgs avatars}
 \centering
\begin{tabular}{lcccc}
\hline
Method         & PSNR $\uparrow$ & SSIM $\uparrow$ & LPIPS $\downarrow$ & FID $\downarrow$ \\ \hline
Ours           & \textbf{30.3607}         & \textbf{0.9682}          & \textbf{0.0339}             & \textbf{33.4665}          \\
3DGS-Avatar \cite{qian20233dgs}   & 28.7836         & 0.9511          & 0.0418             & 49.3673          \\
GaussianAvatar \cite{hu2023gaussianavatar} & 26.9497         & 0.9389          & 0.0407             & 38.5387          \\ \hline
\end{tabular}
\end{table}

%% file: tabs/comparison_relit.tex
 \begin{table}[t]
 \caption{\textbf{Quantitative comparison with state-of-the-art human performance relighting method.}}
 \label{tab: comparison relit}
 \centering
\begin{tabular}{lcccc}
\hline
Method         & PSNR $\uparrow$ & SSIM $\uparrow$ & LPIPS $\downarrow$ & FID $\downarrow$  \\ \hline
Ours           & \textbf{31.6339}         & \textbf{0.9836}           & \textbf{0.0208}    & \textbf{24.1962}                 \\
RA\cite{lin2024relightable} & 24.8013     & 0.9694                 & 0.0738        & 57.3957 \\
R4D\cite{chen2022relighting4d}   & 23.0883       & 0.9608      & 0.0968         & 112.9600                  \\  \hline
\end{tabular}
\end{table}

%% file: tabs/ablation_sparse_view.tex
\begin{table}[t]
\centering
\caption{\textbf{Quantitative evaluation on different view numbers.}}
\label{tab: eval view num}
\footnotesize
\begin{tabular}{lcccc}
\hline
         & PSNR $\uparrow$ & SSIM $\uparrow$ & LPIPS $\downarrow$ & FID $\downarrow$ \\ \hline
3 Views  & \cellcolor{orange!20}30.6123         & \cellcolor{orange!20}0.9807          & \cellcolor{orange!20}0.0306             & 11.3066          \\
6 Views  & 30.3565         & 0.9803          & 0.0310             & \cellcolor{orange!20}10.9966          \\
14 Views & \cellcolor{orange!50}30.7622         & \cellcolor{orange!50}0.9816          & \cellcolor{orange!50}0.0297             & \cellcolor{orange!50}10.6744          \\ \hline
\end{tabular}
\end{table}

%% file: secs/5_discussion.tex
\section{Discussion}

\subsubsection{Conclusion}
We present Animatable Gaussians, a new avatar representation for creating lifelike relightable and animatable human avatars with highly dynamic, realistic and generalized appearances from multi-view RGB videos.
Compared with implicit NeRF-based approaches, we introduce the explicit point-based representation, 3D Gaussian splatting, into the avatar modeling, and leverage powerful 2D CNNs for modeling higher-fidelity human appearances.
Based on the proposed template-guided parameterization and pose projection strategy, our method can not only faithfully reconstruct detailed human appearances, but also generate realistic garment dynamics for novel pose synthesis.
By introducing physically-based rendering into the avatar representation, our method can produce realistic avatar animation under different novel illuminations.
Overall, our method outperforms other state-of-the-art avatar approaches, and we believe that the proposed 3D Gaussian splatting-based avatar representation will make progress towards effective and efficient 3D human representations.

\begin{figure}[t]
    \centering
    \includegraphics[width=\linewidth]{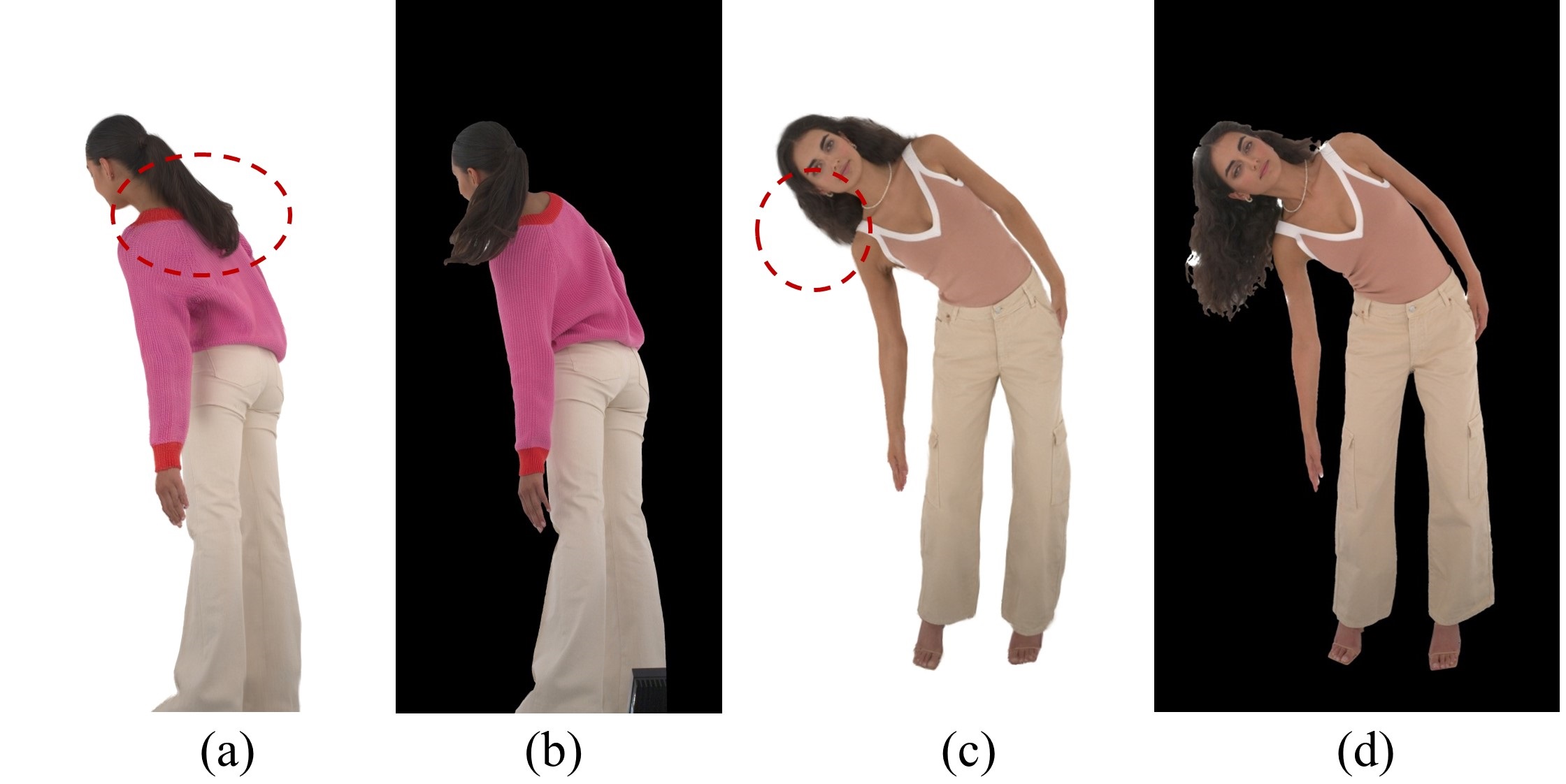}
    \caption{\textbf{Failure cases.}
    (a,c) Animation results by our method, (b,d) ground-truth images.
    Our method fails to model the motion of hairs.
    }
    \label{fig: failure cases}
\end{figure}

\subsubsection{Limitation}
Our method entangles the modeling of the human body and clothes, limiting to changing the clothes of the avatar for applications like virtual try-on.
A possible solution is to separately represent the body and clothes with multi-layer 3D Gaussians as NeRF-based approaches \cite{Feng2022scarf,feng2023learning}.
Moreover, our method relies on the multi-view input to reconstruct a parametric template, limiting the application for modeling loose clothes from a monocular video.
Finally, Our method fails to model the physical motion of components that are not driven by the body joints, e.g., the hairs, as illustrated in Fig.~\ref{fig: failure cases}, since we model the whole body including clothes, hands and hairs as an entangled Gaussian representation.
We leave for future work a disentangled and compositional representation for modeling the dynamics of different components of the character.

%% file: main.bbl
\begin{thebibliography}{100}
\providecommand{\url}[1]{#1}
\csname url@samestyle\endcsname
\providecommand{\newblock}{\relax}
\providecommand{\bibinfo}[2]{#2}
\providecommand{\BIBentrySTDinterwordspacing}{\spaceskip=0pt\relax}
\providecommand{\BIBentryALTinterwordstretchfactor}{4}
\providecommand{\BIBentryALTinterwordspacing}{\spaceskip=\fontdimen2\font plus
\BIBentryALTinterwordstretchfactor\fontdimen3\font minus \fontdimen4\font\relax}
\providecommand{\BIBforeignlanguage}[2]{{%
\expandafter\ifx\csname l@#1\endcsname\relax
\typeout{** WARNING: IEEEtran.bst: No hyphenation pattern has been}%
\typeout{** loaded for the language `#1'. Using the pattern for}%
\typeout{** the default language instead.}%
\else
\language=\csname l@#1\endcsname
\fi
#2}}
\providecommand{\BIBdecl}{\relax}
\BIBdecl

\bibitem{bagautdinov2021driving}
T.~Bagautdinov, C.~Wu, T.~Simon, F.~Prada, T.~Shiratori, S.-E. Wei, W.~Xu, Y.~Sheikh, and J.~Saragih, ``Driving-signal aware full-body avatars,'' \emph{TOG}, vol.~40, no.~4, pp. 1--17, 2021.

\bibitem{xiang2022dressing}
D.~Xiang, T.~Bagautdinov, T.~Stuyck, F.~Prada, J.~Romero, W.~Xu, S.~Saito, J.~Guo, B.~Smith, T.~Shiratori \emph{et~al.}, ``Dressing avatars: Deep photorealistic appearance for physically simulated clothing,'' \emph{TOG}, vol.~41, no.~6, pp. 1--15, 2022.

\bibitem{ma2021power}
Q.~Ma, J.~Yang, S.~Tang, and M.~J. Black, ``The power of points for modeling humans in clothing,'' in \emph{ICCV}, 2021, pp. 10\,974--10\,984.

\bibitem{mildenhall2020nerf}
B.~Mildenhall, P.~P. Srinivasan, M.~Tancik, J.~T. Barron, R.~Ramamoorthi, and R.~Ng, ``Nerf: Representing scenes as neural radiance fields for view synthesis,'' in \emph{ECCV}.\hskip 1em plus 0.5em minus 0.4em\relax Springer, 2020, pp. 405--421.

\bibitem{peng2021animatable}
S.~Peng, J.~Dong, Q.~Wang, S.~Zhang, Q.~Shuai, X.~Zhou, and H.~Bao, ``Animatable neural radiance fields for modeling dynamic human bodies,'' in \emph{ICCV}, 2021, pp. 14\,314--14\,323.

\bibitem{liu2021neural}
L.~Liu, M.~Habermann, V.~Rudnev, K.~Sarkar, J.~Gu, and C.~Theobalt, ``Neural actor: Neural free-view synthesis of human actors with pose control,'' \emph{TOG}, vol.~40, no.~6, pp. 1--16, 2021.

\bibitem{zheng2022structured}
Z.~Zheng, H.~Huang, T.~Yu, H.~Zhang, Y.~Guo, and Y.~Liu, ``Structured local radiance fields for human avatar modeling,'' in \emph{CVPR}, 2022, pp. 15\,893--15\,903.

\bibitem{li2022tava}
R.~Li, J.~Tanke, M.~Vo, M.~Zollh{\"o}fer, J.~Gall, A.~Kanazawa, and C.~Lassner, ``Tava: Template-free animatable volumetric actors,'' in \emph{ECCV}.\hskip 1em plus 0.5em minus 0.4em\relax Springer, 2022, pp. 419--436.

\bibitem{tancik2020fourier}
M.~Tancik, P.~Srinivasan, B.~Mildenhall, S.~Fridovich-Keil, N.~Raghavan, U.~Singhal, R.~Ramamoorthi, J.~Barron, and R.~Ng, ``Fourier features let networks learn high frequency functions in low dimensional domains,'' \emph{NeurIPS}, vol.~33, pp. 7537--7547, 2020.

\bibitem{kerbl2023gaussian}
B.~Kerbl, G.~Kopanas, T.~Leimk{\"u}hler, and G.~Drettakis, ``3d gaussian splatting for real-time radiance field rendering,'' \emph{TOG}, vol.~42, no.~4, pp. 1--14, 2023.

\bibitem{lin2022learning}
S.~Lin, H.~Zhang, Z.~Zheng, R.~Shao, and Y.~Liu, ``Learning implicit templates for point-based clothed human modeling,'' in \emph{ECCV}.\hskip 1em plus 0.5em minus 0.4em\relax Springer, 2022, pp. 210--228.

\bibitem{ma2022neural}
Q.~Ma, J.~Yang, M.~J. Black, and S.~Tang, ``Neural point-based shape modeling of humans in challenging clothing,'' in \emph{3DV}.\hskip 1em plus 0.5em minus 0.4em\relax IEEE, 2022, pp. 679--689.

\bibitem{loper2015smpl}
M.~Loper, N.~Mahmood, J.~Romero, G.~Pons-Moll, and M.~J. Black, ``Smpl: A skinned multi-person linear model,'' \emph{TOG}, vol.~34, no.~6, pp. 1--16, 2015.

\bibitem{karras2019style}
T.~Karras, S.~Laine, and T.~Aila, ``A style-based generator architecture for generative adversarial networks,'' in \emph{CVPR}, 2019, pp. 4401--4410.

\bibitem{karras2020analyzing}
T.~Karras, S.~Laine, M.~Aittala, J.~Hellsten, J.~Lehtinen, and T.~Aila, ``Analyzing and improving the image quality of stylegan,'' in \emph{CVPR}, 2020, pp. 8110--8119.

\bibitem{karras2021alias}
T.~Karras, M.~Aittala, S.~Laine, E.~H{\"a}rk{\"o}nen, J.~Hellsten, J.~Lehtinen, and T.~Aila, ``Alias-free generative adversarial networks,'' \emph{NeurIPS}, vol.~34, pp. 852--863, 2021.

\bibitem{wang2023styleavatar}
L.~Wang, X.~Zhao, J.~Sun, Y.~Zhang, H.~Zhang, T.~Yu, and Y.~Liu, ``Styleavatar: Real-time photo-realistic portrait avatar from a single video,'' in \emph{SIGGRAPH Conference Proceedings}, 2023.

\bibitem{li2024animatablegaussians}
Z.~Li, Z.~Zheng, L.~Wang, and Y.~Liu, ``Animatable gaussians: Learning pose-dependent gaussian maps for high-fidelity human avatar modeling,'' in \emph{CVPR}, 2024.

\bibitem{jin2023tensoir}
H.~Jin, I.~Liu, P.~Xu, X.~Zhang, S.~Han, S.~Bi, X.~Zhou, Z.~Xu, and H.~Su, ``Tensoir: Tensorial inverse rendering,'' in \emph{CVPR}, 2023, pp. 165--174.

\bibitem{gao2023relightable}
J.~Gao, C.~Gu, Y.~Lin, H.~Zhu, X.~Cao, L.~Zhang, and Y.~Yao, ``Relightable 3d gaussian: Real-time point cloud relighting with brdf decomposition and ray tracing,'' \emph{arXiv preprint arXiv:2311.16043}, 2023.

\bibitem{stoll2010video}
C.~Stoll, J.~Gall, E.~De~Aguiar, S.~Thrun, and C.~Theobalt, ``Video-based reconstruction of animatable human characters,'' \emph{TOG}, vol.~29, no.~6, pp. 1--10, 2010.

\bibitem{guan2012drape}
P.~Guan, L.~Reiss, D.~A. Hirshberg, A.~Weiss, and M.~J. Black, ``Drape: Dressing any person,'' \emph{TOG}, vol.~31, no.~4, pp. 1--10, 2012.

\bibitem{xu2011video}
F.~Xu, Y.~Liu, C.~Stoll, J.~Tompkin, G.~Bharaj, Q.~Dai, H.-P. Seidel, J.~Kautz, and C.~Theobalt, ``Video-based characters: creating new human performances from a multi-view video database,'' \emph{TOG}, vol.~30, no.~4, pp. 1--10, 2011.

\bibitem{xiang2021modeling}
D.~Xiang, F.~Prada, T.~Bagautdinov, W.~Xu, Y.~Dong, H.~Wen, J.~Hodgins, and C.~Wu, ``Modeling clothing as a separate layer for an animatable human avatar,'' \emph{TOG}, vol.~40, no.~6, pp. 1--15, 2021.

\bibitem{halimi2022pattern}
O.~Halimi, T.~Stuyck, D.~Xiang, T.~Bagautdinov, H.~Wen, R.~Kimmel, T.~Shiratori, C.~Wu, Y.~Sheikh, and F.~Prada, ``Pattern-based cloth registration and sparse-view animation,'' \emph{TOG}, vol.~41, no.~6, pp. 1--17, 2022.

\bibitem{habermann2021real}
M.~Habermann, L.~Liu, W.~Xu, M.~Zollhoefer, G.~Pons-Moll, and C.~Theobalt, ``Real-time deep dynamic characters,'' \emph{TOG}, vol.~40, no.~4, pp. 1--16, 2021.

\bibitem{habermann2023hdhumans}
M.~Habermann, L.~Liu, W.~Xu, G.~Pons-Moll, M.~Zollhoefer, and C.~Theobalt, ``Hdhumans: A hybrid approach for high-fidelity digital humans,'' \emph{ACM SCA}, vol.~6, no.~3, pp. 1--23, 2023.

\bibitem{sumner2007embedded}
R.~W. Sumner, J.~Schmid, and M.~Pauly, ``Embedded deformation for shape manipulation,'' \emph{TOG}, vol.~26, no.~3, pp. 80--es, 2007.

\bibitem{Kwon2023deliffas}
Y.~Kwon, L.~Liu, H.~Fuchs, M.~Habermann, and C.~Theobalt, ``Deliffas: Deformable light fields for fast avatar synthesis,'' in \emph{NeurIPS}, 2023.

\bibitem{alldieck2018video}
T.~Alldieck, M.~Magnor, W.~Xu, C.~Theobalt, and G.~Pons-Moll, ``Video based reconstruction of 3d people models,'' in \emph{CVPR}, 2018, pp. 8387--8397.

\bibitem{alldieck2018detailed}
------, ``Detailed human avatars from monocular video,'' in \emph{3DV}.\hskip 1em plus 0.5em minus 0.4em\relax IEEE, 2018, pp. 98--109.

\bibitem{zhao2022high}
H.~Zhao, J.~Zhang, Y.-K. Lai, Z.~Zheng, Y.~Xie, Y.~Liu, and K.~Li, ``High-fidelity human avatars from a single rgb camera,'' in \emph{CVPR}, 2022, pp. 15\,904--15\,913.

\bibitem{burov2021dynamic}
A.~Burov, M.~Nie{\ss}ner, and J.~Thies, ``Dynamic surface function networks for clothed human bodies,'' in \emph{ICCV}, 2021, pp. 10\,754--10\,764.

\bibitem{kim2022laplacianfusion}
H.~Kim, H.~Nam, J.~Kim, J.~Park, and S.~Lee, ``Laplacianfusion: Detailed 3d clothed-human body reconstruction,'' \emph{ACM Transactions on Graphics (TOG)}, vol.~41, no.~6, pp. 1--14, 2022.

\bibitem{jiang2022selfrecon}
B.~Jiang, Y.~Hong, H.~Bao, and J.~Zhang, ``Selfrecon: Self reconstruction your digital avatar from monocular video,'' in \emph{CVPR}, 2022, pp. 5605--5615.

\bibitem{xu2023relightable}
Z.~Xu, S.~Peng, C.~Geng, L.~Mou, Z.~Yan, J.~Sun, H.~Bao, and X.~Zhou, ``Relightable and animatable neural avatar from sparse-view video,'' \emph{arXiv preprint arXiv:2308.07903}, 2023.

\bibitem{mihajlovic2021leap}
M.~Mihajlovic, Y.~Zhang, M.~J. Black, and S.~Tang, ``Leap: Learning articulated occupancy of people,'' in \emph{CVPR}, 2021, pp. 10\,461--10\,471.

\bibitem{saito2021scanimate}
S.~Saito, J.~Yang, Q.~Ma, and M.~J. Black, ``Scanimate: Weakly supervised learning of skinned clothed avatar networks,'' in \emph{CVPR}, 2021, pp. 2886--2897.

\bibitem{wang2021metaavatar}
S.~Wang, M.~Mihajlovic, Q.~Ma, A.~Geiger, and S.~Tang, ``Metaavatar: Learning animatable clothed human models from few depth images,'' \emph{NeurIPS}, vol.~34, 2021.

\bibitem{tiwari2021neural}
G.~Tiwari, N.~Sarafianos, T.~Tung, and G.~Pons-Moll, ``Neural-gif: Neural generalized implicit functions for animating people in clothing,'' in \emph{ICCV}, 2021, pp. 11\,708--11\,718.

\bibitem{dong2022pina}
Z.~Dong, C.~Guo, J.~Song, X.~Chen, A.~Geiger, and O.~Hilliges, ``Pina: Learning a personalized implicit neural avatar from a single rgb-d video sequence,'' in \emph{CVPR}, 2022.

\bibitem{ho2023learning}
H.-I. Ho, L.~Xue, J.~Song, and O.~Hilliges, ``Learning locally editable virtual humans,'' in \emph{CVPR}, 2023, pp. 21\,024--21\,035.

\bibitem{deng2020nasa}
B.~Deng, J.~P. Lewis, T.~Jeruzalski, G.~Pons-Moll, G.~Hinton, M.~Norouzi, and A.~Tagliasacchi, ``Nasa neural articulated shape approximation,'' in \emph{ECCV}.\hskip 1em plus 0.5em minus 0.4em\relax Springer, 2020, pp. 612--628.

\bibitem{chen2021snarf}
X.~Chen, Y.~Zheng, M.~J. Black, O.~Hilliges, and A.~Geiger, ``Snarf: Differentiable forward skinning for animating non-rigid neural implicit shapes,'' in \emph{ICCV}, 2021, pp. 11\,594--11\,604.

\bibitem{chen2023fast}
X.~Chen, T.~Jiang, J.~Song, M.~Rietmann, A.~Geiger, M.~J. Black, and O.~Hilliges, ``Fast-snarf: A fast deformer for articulated neural fields,'' \emph{IEEE T-PAMI}, 2023.

\bibitem{mihajlovic2022coap}
M.~Mihajlovic, S.~Saito, A.~Bansal, M.~Zollhoefer, and S.~Tang, ``Coap: Compositional articulated occupancy of people,'' in \emph{CVPR}, 2022, pp. 13\,201--13\,210.

\bibitem{li2022avatarcap}
Z.~Li, Z.~Zheng, H.~Zhang, C.~Ji, and Y.~Liu, ``Avatarcap: Animatable avatar conditioned monocular human volumetric capture,'' in \emph{ECCV}.\hskip 1em plus 0.5em minus 0.4em\relax Springer, 2022, pp. 322--341.

\bibitem{peng2022animatable}
S.~Peng, S.~Zhang, Z.~Xu, C.~Geng, B.~Jiang, H.~Bao, and X.~Zhou, ``Animatable neural implicit surfaces for creating avatars from videos,'' \emph{arXiv preprint arXiv:2203.08133}, 2022.

\bibitem{su2021a-nerf}
S.-Y. Su, F.~Yu, M.~Zollh{\"o}fer, and H.~Rhodin, ``A-nerf: Articulated neural radiance fields for learning human shape, appearance, and pose,'' \emph{NeurIPS}, vol.~34, pp. 12\,278--12\,291, 2021.

\bibitem{Feng2022scarf}
Y.~Feng, J.~Yang, M.~Pollefeys, M.~J. Black, and T.~Bolkart, ``Capturing and animation of body and clothing from monocular video,'' in \emph{SIGGRAPH Asia 2022 Conference Proceedings}, ser. SA '22, 2022.

\bibitem{weng2022humannerf}
C.-Y. Weng, B.~Curless, P.~P. Srinivasan, J.~T. Barron, and I.~Kemelmacher-Shlizerman, ``Humannerf: Free-viewpoint rendering of moving people from monocular video,'' in \emph{CVPR}, 2022, pp. 16\,210--16\,220.

\bibitem{te2022neural}
G.~Te, X.~Li, X.~Li, J.~Wang, W.~Hu, and Y.~Lu, ``Neural capture of animatable 3d human from monocular video,'' in \emph{ECCV}.\hskip 1em plus 0.5em minus 0.4em\relax Springer, 2022, pp. 275--291.

\bibitem{peng2022selfnerf}
B.~Peng, J.~Hu, J.~Zhou, and J.~Zhang, ``Selfnerf: Fast training nerf for human from monocular self-rotating video,'' \emph{arXiv preprint arXiv:2210.01651}, 2022.

\bibitem{guo2023vid2avatar}
C.~Guo, T.~Jiang, X.~Chen, J.~Song, and O.~Hilliges, ``Vid2avatar: 3d avatar reconstruction from videos in the wild via self-supervised scene decomposition,'' in \emph{CVPR}, 2023, pp. 12\,858--12\,868.

\bibitem{jiang2023instantavatar}
T.~Jiang, X.~Chen, J.~Song, and O.~Hilliges, ``Instantavatar: Learning avatars from monocular video in 60 seconds,'' in \emph{CVPR}, 2023, pp. 16\,922--16\,932.

\bibitem{jiang2022neuman}
W.~Jiang, K.~M. Yi, G.~Samei, O.~Tuzel, and A.~Ranjan, ``Neuman: Neural human radiance field from a single video,'' in \emph{ECCV}.\hskip 1em plus 0.5em minus 0.4em\relax Springer, 2022, pp. 402--418.

\bibitem{chen2023uv}
Y.~Chen, X.~Wang, X.~Chen, Q.~Zhang, X.~Li, Y.~Guo, J.~Wang, and F.~Wang, ``Uv volumes for real-time rendering of editable free-view human performance,'' in \emph{CVPR}, 2023, pp. 16\,621--16\,631.

\bibitem{guler2018densepose}
R.~A. G{\"u}ler, N.~Neverova, and I.~Kokkinos, ``Densepose: Dense human pose estimation in the wild,'' in \emph{CVPR}, 2018, pp. 7297--7306.

\bibitem{wang2022arah}
S.~Wang, K.~Schwarz, A.~Geiger, and S.~Tang, ``Arah: Animatable volume rendering of articulated human sdfs,'' in \emph{ECCV}.\hskip 1em plus 0.5em minus 0.4em\relax Springer, 2022, pp. 1--19.

\bibitem{yariv2021volume}
L.~Yariv, J.~Gu, Y.~Kasten, and Y.~Lipman, ``Volume rendering of neural implicit surfaces,'' \emph{NeurIPS}, vol.~34, pp. 4805--4815, 2021.

\bibitem{wang2021neus}
P.~Wang, L.~Liu, Y.~Liu, C.~Theobalt, T.~Komura, and W.~Wang, ``Neus: Learning neural implicit surfaces by volume rendering for multi-view reconstruction,'' in \emph{NeurIPS}, 2021.

\bibitem{su2022danbo}
S.-Y. Su, T.~Bagautdinov, and H.~Rhodin, ``Danbo: Disentangled articulated neural body representations via graph neural networks,'' in \emph{ECCV}.\hskip 1em plus 0.5em minus 0.4em\relax Springer, 2022, pp. 107--124.

\bibitem{li2023posevocab}
Z.~Li, Z.~Zheng, Y.~Liu, B.~Zhou, and Y.~Liu, ``Posevocab: Learning joint-structured pose embeddings for human avatar modeling,'' in \emph{ACM SIGGRAPH Conference Proceedings}, 2023.

\bibitem{dong2022totalselfscan}
J.~Dong, Q.~Fang, Y.~Guo, S.~Peng, Q.~Shuai, X.~Zhou, and H.~Bao, ``Totalselfscan: Learning full-body avatars from self-portrait videos of faces, hands, and bodies,'' in \emph{NeurIPS}, 2022.

\bibitem{shen2023xavatar}
K.~Shen, C.~Guo, M.~Kaufmann, J.~J. Zarate, J.~Valentin, J.~Song, and O.~Hilliges, ``X-avatar: Expressive human avatars,'' in \emph{CVPR}, 2023, pp. 16\,911--16\,921.

\bibitem{zheng2023avatarrex}
Z.~Zheng, X.~Zhao, H.~Zhang, B.~Liu, and Y.~Liu, ``Avatarrex: Real-time expressive full-body avatars,'' \emph{TOG}, vol.~42, no.~4, 2023.

\bibitem{ma2021scale}
Q.~Ma, S.~Saito, J.~Yang, S.~Tang, and M.~J. Black, ``Scale: Modeling clothed humans with a surface codec of articulated local elements,'' in \emph{CVPR}, 2021, pp. 16\,082--16\,093.

\bibitem{zhang2023closet}
H.~Zhang, S.~Lin, R.~Shao, Y.~Zhang, Z.~Zheng, H.~Huang, Y.~Guo, and Y.~Liu, ``Closet: Modeling clothed humans on continuous surface with explicit template decomposition,'' in \emph{CVPR}, 2023.

\bibitem{qi2017pointnet}
C.~R. Qi, H.~Su, K.~Mo, and L.~J. Guibas, ``Pointnet: Deep learning on point sets for 3d classification and segmentation,'' in \emph{CVPR}, 2017, pp. 652--660.

\bibitem{qi2017pointnet++}
C.~R. Qi, L.~Yi, H.~Su, and L.~J. Guibas, ``Pointnet++: Deep hierarchical feature learning on point sets in a metric space,'' in \emph{NeurIPS}, 2017.

\bibitem{prokudin2023dynamic}
S.~Prokudin, Q.~Ma, M.~Raafat, J.~Valentin, and S.~Tang, ``Dynamic point fields,'' in \emph{ICCV}, 2023, pp. 7964--7976.

\bibitem{su2023npc}
S.-Y. Su, T.~Bagautdinov, and H.~Rhodin, ``Npc: Neural point characters from video,'' in \emph{ICCV}, 2023.

\bibitem{xu2022pointnerf}
Q.~Xu, Z.~Xu, J.~Philip, S.~Bi, Z.~Shu, K.~Sunkavalli, and U.~Neumann, ``Point-nerf: Point-based neural radiance fields,'' in \emph{CVPR}, 2022, pp. 5438--5448.

\bibitem{pfister2000surfels}
H.~Pfister, M.~Zwicker, J.~Van~Baar, and M.~Gross, ``Surfels: Surface elements as rendering primitives,'' in \emph{SIGGRAPH}, 2000, pp. 335--342.

\bibitem{zwicker2001surface}
M.~Zwicker, H.~Pfister, J.~Van~Baar, and M.~Gross, ``Surface splatting,'' in \emph{SIGGRAPH}, 2001, pp. 371--378.

\bibitem{zwicker2002pointshop}
M.~Zwicker, M.~Pauly, O.~Knoll, and M.~Gross, ``Pointshop 3d: An interactive system for point-based surface editing,'' \emph{TOG}, vol.~21, no.~3, pp. 322--329, 2002.

\bibitem{zwicker2004perspective}
M.~Zwicker, J.~Rasanen, M.~Botsch, C.~Dachsbacher, and M.~Pauly, ``Perspective accurate splatting,'' in \emph{Proceedings-Graphics Interface}, 2004, pp. 247--254.

\bibitem{yifan2019differentiable}
W.~Yifan, F.~Serena, S.~Wu, C.~{\"O}ztireli, and O.~Sorkine-Hornung, ``Differentiable surface splatting for point-based geometry processing,'' \emph{TOG}, vol.~38, no.~6, pp. 1--14, 2019.

\bibitem{aliev2020neural}
K.-A. Aliev, A.~Sevastopolsky, M.~Kolos, D.~Ulyanov, and V.~Lempitsky, ``Neural point-based graphics,'' in \emph{ECCV}.\hskip 1em plus 0.5em minus 0.4em\relax Springer, 2020, pp. 696--712.

\bibitem{lassner2021pulsar}
C.~Lassner and M.~Zollhofer, ``Pulsar: Efficient sphere-based neural rendering,'' in \emph{CVPR}, 2021, pp. 1440--1449.

\bibitem{kopanas2021point}
G.~Kopanas, J.~Philip, T.~Leimk{\"u}hler, and G.~Drettakis, ``Point-based neural rendering with per-view optimization,'' in \emph{Computer Graphics Forum}, vol.~40, no.~4.\hskip 1em plus 0.5em minus 0.4em\relax Wiley Online Library, 2021, pp. 29--43.

\bibitem{ruckert2022adop}
D.~R{\"u}ckert, L.~Franke, and M.~Stamminger, ``Adop: Approximate differentiable one-pixel point rendering,'' \emph{TOG}, vol.~41, no.~4, pp. 1--14, 2022.

\bibitem{zheng2023pointavatar}
Y.~Zheng, W.~Yifan, G.~Wetzstein, M.~J. Black, and O.~Hilliges, ``Pointavatar: Deformable point-based head avatars from videos,'' in \emph{CVPR}, 2023, pp. 21\,057--21\,067.

\bibitem{ravi2020accelerating}
N.~Ravi, J.~Reizenstein, D.~Novotny, T.~Gordon, W.-Y. Lo, J.~Johnson, and G.~Gkioxari, ``Accelerating 3d deep learning with pytorch3d,'' \emph{arXiv preprint arXiv:2007.08501}, 2020.

\bibitem{luiten2023dynamic}
J.~Luiten, G.~Kopanas, B.~Leibe, and D.~Ramanan, ``Dynamic 3d gaussians: Tracking by persistent dynamic view synthesis,'' \emph{arXiv preprint arXiv:2308.09713}, 2023.

\bibitem{yang2023deformable}
Z.~Yang, X.~Gao, W.~Zhou, S.~Jiao, Y.~Zhang, and X.~Jin, ``Deformable 3d gaussians for high-fidelity monocular dynamic scene reconstruction,'' \emph{arXiv preprint arXiv:2309.13101}, 2023.

\bibitem{wu20234d}
G.~Wu, T.~Yi, J.~Fang, L.~Xie, X.~Zhang, W.~Wei, W.~Liu, Q.~Tian, and X.~Wang, ``4d gaussian splatting for real-time dynamic scene rendering,'' \emph{arXiv preprint arXiv:2310.08528}, 2023.

\bibitem{yang2023real}
Z.~Yang, H.~Yang, Z.~Pan, X.~Zhu, and L.~Zhang, ``Real-time photorealistic dynamic scene representation and rendering with 4d gaussian splatting,'' \emph{arXiv preprint arXiv:2310.10642}, 2023.

\bibitem{zielonka2023drivable}
W.~Zielonka, T.~Bagautdinov, S.~Saito, M.~Zollh{\"o}fer, J.~Thies, and J.~Romero, ``Drivable 3d gaussian avatars,'' \emph{arXiv preprint arXiv:2311.08581}, 2023.

\bibitem{lei2023gart}
J.~Lei, Y.~Wang, G.~Pavlakos, L.~Liu, and K.~Daniilidis, ``Gart: Gaussian articulated template models,'' in \emph{CVPR}, 2024.

\bibitem{qian20233dgs}
Z.~Qian, S.~Wang, M.~Mihajlovic, A.~Geiger, and S.~Tang, ``3dgs-avatar: Animatable avatars via deformable 3d gaussian splatting,'' in \emph{CVPR}, 2024.

\bibitem{hu2023gauhuman}
S.~Hu and Z.~Liu, ``Gauhuman: Articulated gaussian splatting from monocular human videos,'' in \emph{CVPR}, 2024.

\bibitem{kocabas2023hugs}
M.~Kocabas, J.-H.~R. Chang, J.~Gabriel, O.~Tuzel, and A.~Ranjan, ``Hugs: Human gaussian splats,'' in \emph{CVPR}, 2024.

\bibitem{pang2023ash}
H.~Pang, H.~Zhu, A.~Kortylewski, C.~Theobalt, and M.~Habermann, ``Ash: Animatable gaussian splats for efficient and photoreal human rendering,'' in \emph{CVPR}, 2024.

\bibitem{hu2023gaussianavatar}
L.~Hu, H.~Zhang, Y.~Zhang, B.~Zhou, B.~Liu, S.~Zhang, and L.~Nie, ``Gaussianavatar: Towards realistic human avatar modeling from a single video via animatable 3d gaussians,'' in \emph{CVPR}, 2024.

\bibitem{chabert2006relighting}
C.-F. Chabert, P.~Einarsson, A.~Jones, B.~Lamond, W.-C. Ma, S.~Sylwan, T.~Hawkins, and P.~Debevec, ``Relighting human locomotion with flowed reflectance fields,'' in \emph{ACM SIGGRAPH 2006 Sketches}, 2006, pp. 76--es.

\bibitem{debevec2012light}
P.~Debevec, ``The light stages and their applications to photoreal digital actors,'' \emph{SIGGRAPH Asia}, vol.~2, no.~4, pp. 1--6, 2012.

\bibitem{debevec2000acquiring}
P.~Debevec, T.~Hawkins, C.~Tchou, H.-P. Duiker, W.~Sarokin, and M.~Sagar, ``Acquiring the reflectance field of a human face,'' in \emph{Proceedings of the 27th annual conference on Computer graphics and interactive techniques}, 2000, pp. 145--156.

\bibitem{debevec2002lighting}
P.~Debevec, A.~Wenger, C.~Tchou, A.~Gardner, J.~Waese, and T.~Hawkins, ``A lighting reproduction approach to live-action compositing,'' \emph{TOG}, vol.~21, no.~3, pp. 547--556, 2002.

\bibitem{guo2019relightables}
K.~Guo, P.~Lincoln, P.~Davidson, J.~Busch, X.~Yu, M.~Whalen, G.~Harvey, S.~Orts-Escolano, R.~Pandey, J.~Dourgarian \emph{et~al.}, ``The relightables: Volumetric performance capture of humans with realistic relighting,'' \emph{TOG}, vol.~38, no.~6, pp. 1--19, 2019.

\bibitem{hawkins2001photometric}
T.~Hawkins, J.~Cohen, and P.~Debevec, ``A photometric approach to digitizing cultural artifacts,'' in \emph{Proceedings of the 2001 conference on Virtual reality, archeology, and cultural heritage}, 2001, pp. 333--342.

\bibitem{wenger2005performance}
A.~Wenger, A.~Gardner, C.~Tchou, J.~Unger, T.~Hawkins, and P.~Debevec, ``Performance relighting and reflectance transformation with time-multiplexed illumination,'' \emph{TOG}, vol.~24, no.~3, pp. 756--764, 2005.

\bibitem{weyrich2006analysis}
T.~Weyrich, W.~Matusik, H.~Pfister, B.~Bickel, C.~Donner, C.~Tu, J.~McAndless, J.~Lee, A.~Ngan, H.~W. Jensen \emph{et~al.}, ``Analysis of human faces using a measurement-based skin reflectance model,'' \emph{TOG}, vol.~25, no.~3, pp. 1013--1024, 2006.

\bibitem{chen2022relighting4d}
Z.~Chen and Z.~Liu, ``Relighting4d: Neural relightable human from videos,'' in \emph{ECCV}.\hskip 1em plus 0.5em minus 0.4em\relax Springer, 2022, pp. 606--623.

\bibitem{iqbal2023rana}
U.~Iqbal, A.~Caliskan, K.~Nagano, S.~Khamis, P.~Molchanov, and J.~Kautz, ``Rana: Relightable articulated neural avatars,'' in \emph{ICCV}, 2023, pp. 23\,142--23\,153.

\bibitem{sun2023neural}
W.~Sun, Y.~Che, H.~Huang, and Y.~Guo, ``Neural reconstruction of relightable human model from monocular video,'' in \emph{ICCV}, 2023, pp. 397--407.

\bibitem{lin2024relightable}
W.~Lin, C.~Zheng, J.-H. Yong, and F.~Xu, ``Relightable and animatable neural avatars from videos,'' in \emph{AAAI}, vol.~38, no.~4, 2024, pp. 3486--3494.

\bibitem{wang2023intrinsicavatar}
S.~Wang, B.~Anti{\'c}, A.~Geiger, and S.~Tang, ``Intrinsicavatar: Physically based inverse rendering of dynamic humans from monocular videos via explicit ray tracing,'' \emph{arXiv preprint arXiv:2312.05210}, 2023.

\bibitem{peng2021neural}
S.~Peng, Y.~Zhang, Y.~Xu, Q.~Wang, Q.~Shuai, H.~Bao, and X.~Zhou, ``Neural body: Implicit neural representations with structured latent codes for novel view synthesis of dynamic humans,'' in \emph{CVPR}, 2021, pp. 9054--9063.

\bibitem{SMPL-X:2019}
G.~Pavlakos, V.~Choutas, N.~Ghorbani, T.~Bolkart, A.~A.~A. Osman, D.~Tzionas, and M.~J. Black, ``Expressive body capture: 3d hands, face, and body from a single image,'' in \emph{CVPR}, 2019.

\bibitem{lorensen1987marching}
W.~E. Lorensen and H.~E. Cline, ``Marching cubes: A high resolution 3d surface construction algorithm,'' \emph{TOG}, vol.~21, no.~4, pp. 163--169, 1987.

\bibitem{kajiya1986rendering}
J.~T. Kajiya, ``The rendering equation,'' in \emph{Proceedings of the 13th annual conference on Computer graphics and interactive techniques}, 1986, pp. 143--150.

\bibitem{zhang2018unreasonable}
R.~Zhang, P.~Isola, A.~A. Efros, E.~Shechtman, and O.~Wang, ``The unreasonable effectiveness of deep features as a perceptual metric,'' in \emph{CVPR}, 2018, pp. 586--595.

\bibitem{saito2020pifuhd}
S.~Saito, T.~Simon, J.~Saragih, and H.~Joo, ``Pifuhd: Multi-level pixel-aligned implicit function for high-resolution 3d human digitization,'' in \emph{CVPR}, June 2020.

\bibitem{deng2024ramavatar}
X.~Deng, Z.~Zheng, Y.~Zhang, J.~Sun, C.~Xu, x.~Yang, L.~Wang, and Y.~Liu, ``Ram-avatar: Real-time photo-realistic avatar from monocular videos with full-body control,'' in \emph{CVPR}, 2024.

\bibitem{mahmood2019amass}
N.~Mahmood, N.~Ghorbani, N.~F. Troje, G.~Pons-Moll, and M.~J. Black, ``Amass: Archive of motion capture as surface shapes,'' in \emph{ICCV}, 2019, pp. 5442--5451.

\bibitem{isik2023humanrf}
M.~I\c{s}{\i}k, M.~Rünz, M.~Georgopoulos, T.~Khakhulin, J.~Starck, L.~Agapito, and M.~Nießner, ``Humanrf: High-fidelity neural radiance fields for humans in motion,'' \emph{TOG}, vol.~42, no.~4, pp. 1--12, 2023.

\bibitem{zhang2021lightweight}
Y.~Zhang, Z.~Li, L.~An, M.~Li, T.~Yu, and Y.~Liu, ``Lightweight multi-person total motion capture using sparse multi-view cameras,'' in \emph{ICCV}, 2021, pp. 5560--5569.

\bibitem{wang2004image}
Z.~Wang, A.~C. Bovik, H.~R. Sheikh, and E.~P. Simoncelli, ``Image quality assessment: from error visibility to structural similarity,'' \emph{IEEE T-IP}, vol.~13, no.~4, pp. 600--612, 2004.

\bibitem{heusel2017gans}
M.~Heusel, H.~Ramsauer, T.~Unterthiner, B.~Nessler, and S.~Hochreiter, ``Gans trained by a two time-scale update rule converge to a local nash equilibrium,'' \emph{NeurIPS}, vol.~30, 2017.

\bibitem{broyden1965class}
C.~G. Broyden, ``A class of methods for solving nonlinear simultaneous equations,'' \emph{Mathematics of computation}, vol.~19, no.~92, pp. 577--593, 1965.

\bibitem{gropp2020implicit}
A.~Gropp, L.~Yariv, N.~Haim, M.~Atzmon, and Y.~Lipman, ``Implicit geometric regularization for learning shapes,'' in \emph{ICML}.\hskip 1em plus 0.5em minus 0.4em\relax PMLR, 2020, pp. 3789--3799.

\bibitem{adam}
D.~P. Kingma and J.~Ba, ``Adam: A method for stochastic optimization,'' in \emph{ICLR}, 2015.

\bibitem{ronneberger2015u}
O.~Ronneberger, P.~Fischer, and T.~Brox, ``U-net: Convolutional networks for biomedical image segmentation,'' in \emph{MICCAI}.\hskip 1em plus 0.5em minus 0.4em\relax Springer, 2015, pp. 234--241.

\bibitem{feng2023learning}
Y.~Feng, W.~Liu, T.~Bolkart, J.~Yang, M.~Pollefeys, and M.~J. Black, ``Learning disentangled avatars with hybrid 3d representations,'' \emph{arXiv preprint arXiv:2309.06441}, 2023.

\end{thebibliography}
